\documentclass[10pt,twocolumn,letterpaper]{article}
\usepackage{cvpr}
\usepackage[dvipsnames]{xcolor}
\usepackage{xspace}
\definecolor{cvprblue}{rgb}{0.21,0.49,0.74}
\usepackage[pagebackref,breaklinks,colorlinks,citecolor=cvprblue]{hyperref}
\usepackage{bbm}

\newcommand{\method}{Splatter Image\xspace}

\makeatletter
\renewcommand{\paragraph}{%
    \@startsection{paragraph}{4}%
    {\z@}{-0.5em}{-0.5em}%
    {\normalfont\normalsize\bfseries}%
}
\makeatother

\usepackage{pifont}
\newcommand{\xmark}{\ding{55}}%

\title{\method: Ultra-Fast Single-View 3D Reconstruction}

\author{Stanislaw Szymanowicz \quad Christian Rupprecht \quad Andrea Vedaldi\\[0.3em]
Visual Geometry Group --- University of Oxford\\
{\tt\small \{stan,chrisr,vedaldi\}@robots.ox.ac.uk}
\vspace{-1em}%
}

\begin{document}

\twocolumn[{%
\maketitle
\thispagestyle{empty}
\begin{center}
\centering
\captionsetup{type=figure}
\includegraphics[width=\textwidth]{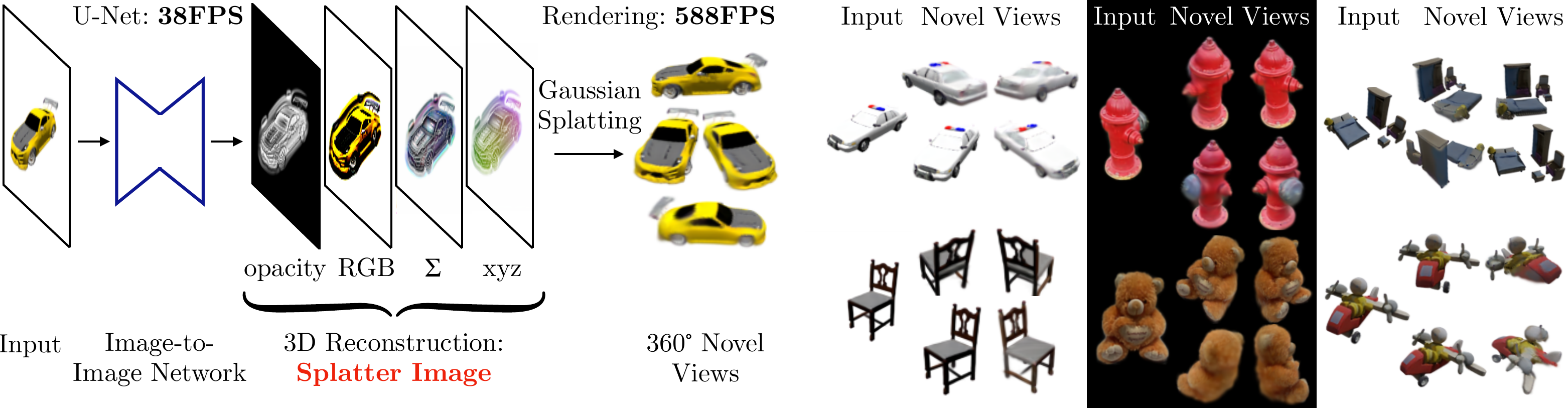}
\captionof{figure}{
The \textbf{\method} is an ultra-efficient method for single- and few-view 3D reconstruction.
It uses an image-to-image neural network to map the input image to another image that holds the parameters of one coloured 3D Gaussian per pixel.
Splatter Image achieves excellent 3D reconstruction quality on synthetic, real and large-scale datasets while using a single GPU for training.
}%
\label{fig:teaser-fig}%
\end{center}%
\vspace{1em}%
}]

\begin{abstract}
We introduce the \method, an ultra-efficient approach for monocular 3D object reconstruction.
\method is based on Gaussian Splatting, which allows fast and high-quality reconstruction of 3D scenes from multiple images.
We apply Gaussian Splatting to  monocular reconstruction by learning a neural network that, at test time, performs reconstruction in a feed-forward manner, at 38 FPS\@.
Our main innovation is the surprisingly straightforward design of this network, which, using 2D operators, maps the input image to one 3D Gaussian per pixel.
The resulting set of Gaussians thus has the form an image, the \method.
We further extend the method take several images as input via cross-view attention.
Owning to the speed of the renderer (588 FPS), we use a single GPU for training while generating entire images at each iteration to optimize perceptual metrics like LPIPS\@.
On several synthetic, real, multi-category and large-scale benchmark datasets, we achieve better results in terms of PSNR, LPIPS, and other metrics while training and evaluating much faster than prior works.
Code, models, demo and more results are available at {\url{https://szymanowiczs.github.io/splatter-image}}.
\end{abstract}
\section{Introduction}%
\label{sec:intro}

We contribute \method, a method that achieves ultra-fast single-view reconstruction of the 3D shape and appearance of objects.
\method uses a set of 3D Gaussians as the 3D representation, taking advantage of the rendering quality and speed of Gaussian Splatting~\cite{kerbl233d-gaussian}.
\method works by predicting a 3D Gaussian for each of the input image pixels, using an image-to-image neural network.
Remarkably, the predicted 3D Gaussians provide 360$^{\circ}$ reconstructions of quality comparable or superior to much slower methods (\cref{fig:teaser-fig}).

We formulate monocular 3D reconstruction as the problem of designing a neural network that takes an image of an object as input and produces as output a corresponding Gaussian mixture that represents all sides of it.
While a Gaussian mixture is a set, \ie, an unordered collection, it can still be stored in an ordered data structure.
\method takes advantage of this fact by using a \emph{2D image} as the container of the 3D Gaussians, storing the parameters of one Gaussian (\ie, its opacity, position, shape, and colour) per pixel.
The Gaussians predominantly lie on the rays from the camera to the object, but they can also be placed off the rays (\cref{fig:method/location}), enabling 360$^{\circ}$ object representation.

The advantage of storing a set of 3D Gaussians in an image is that it reduces the reconstruction problem to learning an image-to-image neural network.
In this manner, the reconstructor can be implemented utilizing only efficient 2D operators (\eg, 2D convolution instead of 3D convolution).
We use in particular a U-Net~\cite{ronneberger15u-net:} as those have demonstrated excellent performance in image generation~\cite{rombach22high-resolution}.
In our case, their ability to capture small image details~\cite{yu21pixelnerf:} helps to obtain higher-quality reconstructions.

Since the 3D representation in \method is a mixture of 3D Gaussians, it enjoys the rendering and space efficiency of Gaussian Splatting, which benefits inference and training.
In particular, rendering stops being a training bottleneck~\cite{mildenhall20nerf:} and we can afford to generate complete views of the object to optimize perceptual metrics like LPIPS~\cite{zhang2018perceptual}.
More importantly, the efficiency is such that our model can be trained on a \emph{single GPU} on standard benchmarks of 3D objects or two GPUs on large datasets such as Objaverse~\cite{deitke23objaverse:}, whereas alternative methods typically require distributed training on dozens~\cite{lin2023visionnerf} or even hundreds~\cite{hong24lrm:} of GPUs.
We also extend \method to take several views as input.
This is achieved by taking the union of the Gaussian mixtures predicted from individual views, after registering them to a common coordinate frame.
The different views communicate during prediction via lightweight cross-view attention layers in the architecture.

Empirically, we show that, while the network only sees one side of the object, it can still produce a 360$^\circ$ reconstruction of it by using the prior acquired during training.
The 360$^\circ$ information is encoded in the 2D image by allocating different Gaussians in a given 2D neighbourhood to different parts of the 3D object.

We validate \method by comparing it to alternative, slower reconstructors on standard benchmark datasets like ShapeNet~\cite{chang15shapenet} and CO3D~\cite{reizenstein21co3d}.
To assess scalability and generalization, we also apply \method to multi-category reconstruction and train it on Objaverse~\cite{deitke23objaverse:}, and evaluate it on Google Scanned Objects~\cite{francis22gso}.
We obtain results of quality comparable to the recent Large Reconstruction Model of~\cite{he2023openlrm,hong24lrm:}, which is $50 \times$ more expensive to train.
In fact, in several cases we even outperform slower methods in PSNR and LPIPS\@.
We argue that this is because the very efficient design allows training the model very effectively, including using image-level losses like LPIPS\@.

To summarise, our contributions are:
(1) to port Gaussian Splatting to learning-based monocular reconstruction;
(2) to do so with the \method, a straightforward, efficient and performant 3D reconstruction approach that operates at 38 FPS on a standard GPU\@ and affords single-GPU training;
(3) to also extend the method to multi-view reconstruction;
(4) and to obtain state-of-the-art reconstruction performance in multiple standard benchmarks, including synthetic, real, multi-category and large-scale datasets, in terms of reconstruction quality and speed.

\section{Related work}%
\label{s:related}

\paragraph{Representations for single-view 3D reconstruction.}

In recent years, implicit representations like NeRF~\cite{mildenhall20nerf:} have dominated learning-based few-view reconstruction, parameterising the MLP in NeRF using global~\cite{jang21codenerf:,rematas21sharf:}, local~\cite{yu21pixelnerf:} or both global and local codes~\cite{lin2023visionnerf}.
However, implicit representations, particularly MLP-based ones, are notoriously slow to render, up to 2s for a single $128 \times 128 $ image.

Follow-up works~\cite{guo2022fenvs,szymanowicz2023viewset} used faster, explicit, voxel grid representations that encode opacities and colours directly.
Similar to DVGO~\cite{sun22direct}, they achieve significant speed-ups, but, due to their voxel-based representation, they scale poorly with resolution.
They also assume the knowledge of the absolute viewpoint of each object image.

The triplane representation~\cite{chan22efficient,chen22tensorf:} was proposed as a compromise between rendering speed and memory consumption.
While they are not as fast to render as explicit representations, they allow view-space reconstruction~\cite{gu2023nerfdiff} and are fast enough to be effectively used for single-view reconstruction~\cite{anciukevicius22renderdiffusion:,gu2023nerfdiff}.
Triplane-based reconstructors were shown to scale to large datasets like Objaverse~\cite{deitke23objaverse:,deitke23objaverse-xl:}, albeit at the cost of hundreds of GPUs for multiple days~\cite{hong24lrm:,tochilkin24triposr:}.

In contrast to these works, our method predicts a mixture of 3D Gaussians in a feed-forward manner.
As a result, our method is cheap to train (1-2 GPUs), fast at inference and achieves real-time rendering speeds while achieving state-of-the-art image quality across multiple metrics on multiple standard single-view reconstruction benchmarks, including single-~\cite{sitzmann2019srns} and multi-category ShapeNet~\cite{chang15shapenet,kato2018renderer}.

When more than one view is available at the input, one can use them to estimate the scene geometry~\cite{chen21mvsnerf:,long22sparseneus:}, learn a view interpolation function~\cite{wang21ibrnet:} or optimize a 3D representation of a scene using priors~\cite{jain2021dietnerf}.
Our method is primarily a single-view reconstruction network, but we do show how \method can be extended to fuse multiple views.
However, we focus our work on object-centric reconstruction rather than on generalising to unseen scenes.

\paragraph{3D Reconstruction with Point Clouds.}

PointOutNet~\cite{fan17point_set} took image encoding as input and trained point cloud prediction networks~\cite{qi2017pointnet} using 3D point cloud supervision.
PVD~\cite{zhou21pvd} and PC$^{2}$~\cite{melas2023pc2} extended this approach using Diffusion Models~\cite{ho20denoising} by conditioning the denoising process on partial point clouds and RGB images, respectively.
These approaches require ground truth 3D point clouds, limiting their applicability.
Other works~\cite{liu21infinite,rockwell21pixelsynth,wiles20synsin} use point clouds as intermediate 3D representations for conditioning 2D inpainting or generation networks.
However, these point clouds are assumed to correspond to only visible object points.
In contrast, our Gaussians can model any part of the object, and thus afford 360$^\circ$ reconstruction.

Point cloud-based representations have also been used for high-quality reconstruction from multi-view images.
Novel views can be rendered with 2D inpainting networks for hole-filling~\cite{ruckert22adop}, or by using non-isotropic 3D Gaussians with variable scale~\cite{kerbl233d-gaussian}.
While showing high-quality results, Gaussian Splatting~\cite{kerbl233d-gaussian} requires many images per scene and has not yet been used in a learning-based reconstruction framework as we do here.

Our method also uses 3D Gaussians as an underlying representation but predicts them from as few as a single image.
Moreover, it outputs a full 360$^\circ$ 3D reconstruction without using 2D or 3D inpainting networks.

\paragraph{Probabilistic 3D Reconstruction.}

Single-view 3D reconstruction is an ambiguous problem, so recently it has been tackled as a conditional generation task.
Diffusion Models have been employed for conditional novel view synthesis~\cite{chan2023genvs,watson20223dim,liu23zero-1-to-3:}.
Due to generating images without underlying geometries, the output images exhibit noticeable flicker.
This can be mitigated by simultaneously generating multi-view images~\cite{liu2023syncdreamer,shi2023MVDream}, reconstructing a geometry at every step of the denoising process~\cite{szymanowicz2023viewset,tewari2023diffusion,xu24dmv3d:} or training a robust reconstructor~\cite{liu23one-2-3-45:,li24instant3d:}.
Other works build and use a 3D~\cite{muller23diffrf:,chen23ssdnerf} or 2D~\cite{gu2023nerfdiff,melas2023realfusion} prior which can be used in an image-conditioned auto-decoding framework.

Here, we focus on deterministic reconstruction.
However, few-view reconstruction is required to output 3D geometries from feed-forward methods~\cite{liu2023syncdreamer,shi2023MVDream,szymanowicz2023viewset,tewari2023diffusion,xu24dmv3d:}.
Our method is capable of few-view 3D reconstruction, thus it is complementary to these generative methods and could lead to improvements in generation speed and quality.

\section{Method}%
\label{s:method}

\newcommand{\x}{\boldsymbol{x}}
\newcommand{\y}{\boldsymbol{y}}
\newcommand{\bbb}{\boldsymbol{b}}
\newcommand{\bu}{\boldsymbol{u}}
\newcommand{\bc}{\boldsymbol{c}}
\newcommand{\bv}{\boldsymbol{v}}
\newcommand{\bq}{\boldsymbol{q}}
\newcommand{\bs}{\boldsymbol{s}}
\newcommand{\bnu}{\boldsymbol{\nu}}
\newcommand{\balpha}{\boldsymbol{\alpha}}
\newcommand{\bmu}{\boldsymbol{\mu}}

We discuss Gaussian Splatting in~\cref{s:gaussian-splatting} for background, and then describe the \method in~\cref{s:image,s:learning,s:multiple,s:color,s:network}.

\subsection{Overview of Gaussian Splatting}%
\label{s:gaussian-splatting}

A \emph{radiance field}~\cite{mildenhall20nerf:} is a pair of functions, assigning an opacity $\sigma(\x)\in\mathbb{R}_+$ and a colour $c(\x,\bnu) \in \mathbb{R}^3$ to each 3D point $\x\in\mathbb{R}^3$ and viewing direction $\bnu\in\mathbb{S}^2$.
Gaussian Splatting~\cite{zwicker01ewa-volume} represents the two functions $\sigma$ and $c$ as a mixture $\theta$ of $G$ colored 3D Gaussians
$$
g_i(\x) =
\exp \left( -\frac{1}{2} (\x - \bmu_i)^\top \Sigma_i^{-1} (\x - \bmu_i) \right),
$$
where $1 \leq i \leq G$, $\bmu_i \in \mathbb{R}^3$ is the Gaussian mean or center and $\Sigma_i \in \mathbb{R}^{3\times3}$ is its covariance, specifying its shape and size.
Each Gaussian has also an opacity $\sigma_i \in [0, 1]$ and a view-dependent colour $c_i(\bv) \in \mathbb{R}^3$.
Together, they define a radiance field as follows:
\begin{equation}\label{e:rf}
\sigma(\x)
=
\sum_{i=1}^G
\sigma_i g_i(\x),
~~~~
c(\x,\bnu)
=
\frac{ \sum_{i=1}^G c_i(\bnu) \sigma_i g_i(\x)}{ \sum_{j=1}^G \sigma_i g_i(\x)}.
\end{equation}
The mixture of Gaussians is thus given by the \emph{set}
$$
\theta= \{(\sigma_i, \bmu_i,\Sigma_i,c_i), i=1,\dots, G\}.
$$

A raidance field is rendered into an image $I(\bu)$ by integrating the colors observed along the ray $\x_{\tau} = \x_0 - \tau \bnu$,
$\tau \in \mathbb{R}_+$ that passes through each image pixel $\bu$ via the equation:
\begin{equation}\label{e:ea}
I(\bu)
=
\int_{0}^\infty c(\x_\tau,\bnu) \sigma(\x_\tau)
e^{
  - \int_{0}^\tau \sigma(\x_\mu) \,d\mu
}\,d\tau.
\end{equation}
Gaussian Splatting~\cite{zwicker01ewa-volume,kerbl233d-gaussian} provides a very fast differentiable renderer
$
I = \mathcal{R}(\theta, \pi)
$
that approximates \cref{e:ea}, mapping the mixture $\theta$ and the viewpoint $\pi$ to an image $I$.

\subsection{The \method}%
\label{s:image}

To perform monocular reconstruction we seek for a function $\theta = \mathcal{S}(I)$ which is the `inverse' of the renderer $\mathcal{R}$, mapping an image $I$ to a mixture of 3D Gaussians $\theta$.
Our key innovation is to propose an extremely simple and yet effective design for such a function.
Specifically, we predict a Gaussian for each pixel of the input image $I$ using a standard image-to-image neural network architecture.
We call its output image $M$ the \method.

In more detail, Let $\bu = (u_1,u_2,1)$ denote one of the $H \times W$ image pixels.
This corresponds to ray $\x = \bu d$ in camera space, where $d$ is the depth of the ray point.
Our network $f$ takes as input the $H\times W\times 3$ RGB image $I$, and outputs directly a $H\times W\times K$ tensor $M$, where each pixel is associated to the $K$-dimensional feature vector packing the parameters $M_{\bu} = (\sigma,\bmu,\Sigma,c)$ of a corresponding Gaussian.

\begin{figure}
\centering
\includegraphics[width=0.99\columnwidth]{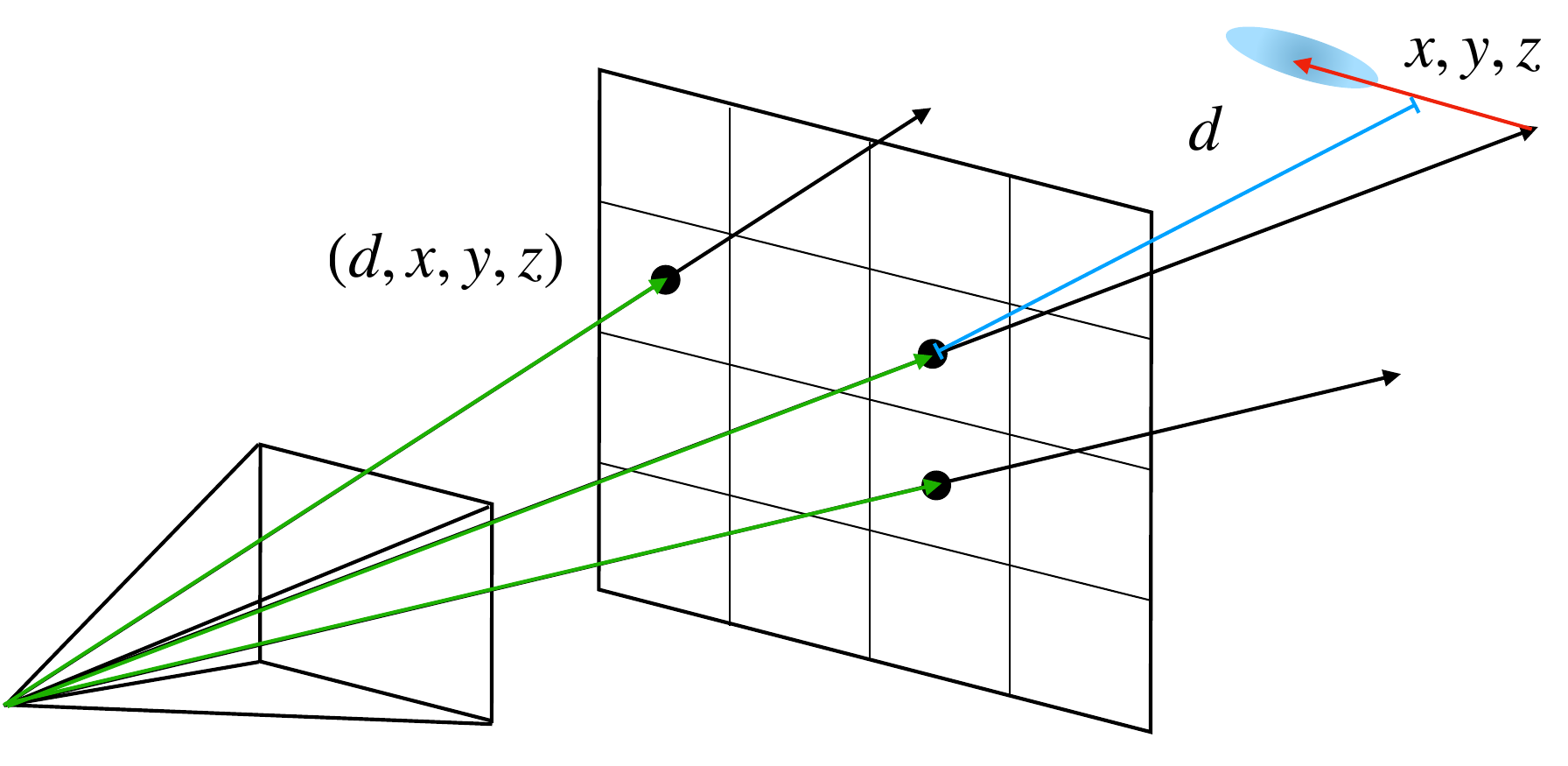}
\caption{\textbf{Predicting locations.} The location of each Gaussian is parameterised by depth $d$ and a 3D offset $\Delta=(\Delta_x, \Delta_y, \Delta_z$). The 3D Gaussians are projected to depth $d$ (blue) along camera rays (green) and moved by the 3D offset $\Delta$ (red).}%
\label{fig:method/location}
\end{figure}

We assume that Gaussians are expressed in the same reference frame of the camera.
As illustrated in \cref{fig:method/location}, the network predicts the depth $d$ and offset $(\Delta_x,\Delta_y,\Delta_y)$, setting
\begin{equation}\label{e:mu}
\bmu = \begin{bmatrix}
  u_1 d + \Delta_x \\
  u_2 d + \Delta_y \\
  d + \Delta_z
\end{bmatrix}.
\end{equation}
The network also predicts the opacity $\sigma$, the shape $\Sigma$ and the colour $c$.
For now, we assume that the colour is Lambertian, \ie,
$c(\nu)= c \in \mathbb{R}^3$, and relax this assumption in \cref{s:color}.
\Cref{s:network} provides more detail on the network architecture.

\paragraph{Discussion.}

One may wonder how this design can predict a full 360$^\circ$ reconstruction of the object when the reconstruction is aligned a single input view.
We find that the network adjusts the 3D offsets $\Delta$ and depths $d$ to allocate some of the 3D Gaussians to reconstruct the input view, and some to reconstruct unseen portions of the object, automatically.
The network can also decide to switch off any Gaussian by simply predicting $\sigma=0$, if needed.
These points are then not rendered and can be culled in post-processing.

Our design can also be seen as an extension of depth prediction networks which only predicting the depth of each pixel.
Here, we also predict unobserved parts of the geometry, as well as the shape and appearance of each Gaussian.

\subsection{Learning formulation}%
\label{s:learning}

Learning to predict the \method is simple and efficient.
It can be done on a single GPU using at most 20GB of memory at training time in most of our single-view reconstruction experiments (except for Objaverse, where we use 2 GPUs and using 26GB of memory on each).
For training, we assume a multi-view dataset, either real or synthetic.
The dataset $\mathcal{D}$ consists of triplets $(I,J,\pi)$, where $I$ is a source image, $J$ a target image, and $\pi$ the viewpoint change between the source and the target cameras.
Then we simply feed the source $I$ as input to \method, and minimize the average reconstruction loss of target view $J$:
\begin{equation}\label{e:loss}
\mathcal{L}(\mathcal{S})
=
\frac{1}{|\mathcal{D}|}
\sum_{(I,J,\pi) \in \mathcal{D}}
\|
J - \mathcal{R}(\mathcal{S}(I), \pi)
\|^2.
\end{equation}

\paragraph{Image-level losses.}

A main advantage of the speed and efficiency of our method is that it allows for rendering entire images at each training iteration, even for relatively large batches (this differs from NeRF~\cite{mildenhall20nerf:}, which only generates a certain number of pixels in a batch).
In particular, this means that, in addition to decomposable losses like the $L2$ loss above, we can use image-level losses like LPIPS~\cite{zhang2018perceptual}, which do not decompose into per-pixel losses.
In practice, we experiment with a combination of such losses.

\paragraph{Regularisations.}

We also add generic regularisers to prevent parameters from taking on unreasonable values (\eg, Gaussians which are larger than the reconstructed objects, or vanishingly small).
Please see the sup.~mat.~for details.

\subsection{Extension to multiple input viewpoints}%
\label{s:multiple}

If two or more input views $I_j$, $j \in \{1,\dots,N\}$ are provided, we can apply network $\mathcal{S}$ multiple times to obtain multiple Splatter Images $M_j$, one per view.
If $(R,T)$ is the relative camera pose change from an additional view to the reference view, we can take the mixture of 3D Gaussians $\theta$ defined in the additional view's coordinates and warp it to the reference view.
Specifically, a Gaussian of parameters
$
(\sigma,\bmu,\Sigma,c)
$
maps to Gaussian of parameters
$
(\sigma,\tilde \bmu, \tilde\Sigma, \tilde c)
$
where
$
\tilde \bmu = R \bmu + T
$,
$
\tilde \Sigma = R \Sigma R^\top
$,
$
\tilde c = c
$.
We use the symbol $\phi[\theta]$ to denote the Gaussian mixture obtained by warping each Gaussian in $\theta$.
Here we have also assumed a Lambertian colour model and will discuss in \cref{s:color} how more complex models transform.

Given $N$ different views $I_j$ and corresponding warps $\phi$, we can obtain a composite mixture of 3D Gaussians simply by taking their union
$
\Theta = \bigcup_{j=1}^N \phi_j[\mathcal{S}(I_j)].
$
Note that this set of 3D Gaussians is defined in the coordinate system of the reference camera.

\subsection{View-dependent colour}%
\label{s:color}

Generalising beyond the Lambertian colour model, we use \emph{spherical harmonics}~\cite{kerbl233d-gaussian} to represent view-dependent colours.
For a particular Gaussian $(\sigma,\bmu,\Sigma,c)$, we then define
$
[c(\bnu;\balpha)]_i
=
\sum_{l=0}^L
\sum_{m=-L}^{L}
\alpha_{ilm} Y_l^m(\bnu)
$
where $\alpha_{ilm}$ are coefficients predicted by the network and $Y_l^m$ are spherical harmonics, $L$ is the order of the expansion, and $\bnu\in\mathbb{S}^2$ is the viewing direction.

The viewpoint change of~\cref{s:multiple} transforms a viewing direction $\bnu$ in the source camera to the corresponding viewing direction in the reference frame as $\tilde\bnu = R \bnu$.
We can then find the transformed colour function by finding the coefficients  $\tilde \balpha$ such that
$
c(\bnu;\balpha) = c(\tilde\bnu;\tilde\balpha).
$
This is possible because (each order of) spherical harmonics are closed under rotation.
However, the general case requires the computation of Wigner matrices.
For simplicity, we only consider orders $L=0$ (Lambertian) and $L=1$.
Hence, the first level has one constant component $Y_0^0$ and the second level has three components which we can write collectively as $Y_1 = [Y_1^{-1}, Y_1^0, Y_1^1]$ such that
$$
Y_1(\bnu) = \sqrt{\frac{3}{4 \pi}} \Pi \bnu,
\quad
\Pi =
\begin{bmatrix}
  0 & 1 & 0 \\
  0 & 0 & 1 \\
  1 & 0 & 0 \\
\end{bmatrix}.
$$
We can then conveniently rewrite
$
[c(\bnu;\balpha)]_i = \alpha_{i0} + \balpha_{i1}^\top Y_1(\bnu).
$
From this and
$
c(\bnu;\alpha_0,\balpha_1)
=
c(\tilde\bnu;\tilde\alpha_0,\tilde\balpha_1)
$
we conclude that
$
\tilde \alpha_{i0} =
\tilde \alpha_{i0}
$,
and
$
\tilde \balpha_{i1} = \Pi^{-1} R \Pi \balpha_{i1}.
$

\subsection{Neural network architecture}%
\label{s:network}

The bulk of the predictor $\mathcal{S}$ mapping the input image $I$ to the mixture of Gaussians $\theta$ is architecturally identical to the SongUNet of~\cite{song2020denoising}.
The last layer is replaced with a $1\times 1$ convolutional layer with $12+k_c$ output channels, where $k_c \in \{3,12\}$ depending on the colour model.
Given $I \in \mathbb{R}^{3\times H\times W}$ as input, the network thus produces a $(12+k_c)\times H\times W$ tensor as output, coding, for each pixel $\bu$ channels, the parameters $(\hat \sigma, \Delta, \hat d, \hat \bs, \hat{\bq}, \balpha)$ which are then transformed to opacity, offset, depth, scale, rotation and colour, respectively.
These are activated by non-linear functions to obtain the Gaussian parameters.
Specifically, the opacity is obtained using the sigmoid operator as $\sigma = \operatorname{sigmoid}(\hat \sigma)$.
The depth is obtained as $d = (z_\text{far} -  z_\text{near}) \operatorname{sigmoid}(\hat d) + z_\text{near}$.
The mean $\bmu$ is then obtained using~\cref{e:mu}.
Following~\cite{kerbl233d-gaussian}, the covariance is obtained as $\Sigma = R(\bq) \operatorname{diag}(\exp{\hat \bs})^2 R(\bq)^\top$ where $R(\bq)$ is the rotation matrix with quaternion $\bq = \hat \bq / \| \hat \bq \|$ and $\hat\bq\in\mathbb{R}^4$.

For multi-view reconstruction, we apply the same network to each input view and then use the approach of~\cref{s:multiple} to fuse the individual reconstructions.
In order to allow the network to coordinate and exchange information between views, we apply two modifications to it.

First, we condition the network with the corresponding camera pose  $(R,T)$ (we only assume access to the \emph{relative} camera pose to a common but otherwise arbitrary reference frame).
In fact, since we consider cameras in a turn-table-like configuration, we only pass vectors $(R\boldsymbol{e}_3,T)$ where $\boldsymbol{e}_3=(0,0,1)$.
We do so by encoding each entry via a sinusoidal positional embedding of order 9, resulting in 60 dimensions in total.
Finally, these are applied to the U-Net blocks via FiLM~\cite{perez18film:} embeddings.

Second, we add cross-attention layers to allow communication between the features of different views.
We do so in a manner similar to~\cite{shi2023MVDream}, but only at the lowest UNet resolution, which maintains the computational cost very low.

\section{Experiments}%
\label{sec:experiments}

We evaluate our method extensively for single-view reconstruction on six standard benchmarks.
Next, we assess the quality of multi-view reconstruction, and finish with an evaluation of the speed of the method.

\paragraph{Datasets.}

The standard benchmark for evaluating single-view 3D reconstruction is ShapeNet-SRN~\cite{sitzmann2019srns}.
We train our method in the single-class setting and report results on the ``Car'' and ``Chair'' classes, following prior work.
Moreover, we challenge our method with two classes of real objects from the CO3D~\cite{reizenstein21co3d} dataset: Hydrants and Teddybears.
In this challenging dataset ripe with ambiguities we set $z_{\text{far}}$ and $z_{\text{near}}$ to depend on ground truth distance $z_{\text{gt}}$ between the object and camera.

We further test our method on two multi-category datasets.
First, we use the standard benchmark of multi-category ShapeNet (with objects from 13 largest categories), and use the renderings, standard splits and target views from NMR~\cite{kato2018renderer}.
Secondly, we train one model on renderings of objects from Objaverse-LVIS~\cite{deitke23objaverse:} which contains over 1k object categories, using the renderings from Zero-1-to-3~\cite{liu23zero-1-to-3:}.
We evaluate this model on all objects from the Google Scanned Objects dataset~\cite{francis22gso}, using the same renderings as used for evaluation in Free3D~\cite{zheng24free3d}.
We train and evaluate all models at $128 \times 128$ resolution, apart from multi-category ShapeNet which is at $64 \times 64$.
Finally, we use the ShapeNet-SRN Cars dataset for the evaluation of the two-view reconstruction quality.
For more details on datasets see supp.~mat.

\paragraph{Baselines.}

For ShapeNet (both single-class and multi-class), we compare against implicit~\cite{jang21codenerf:,lin2023visionnerf,sitzmann2019srns,yu21pixelnerf:}, hydrid implicit-explicit~\cite{gu2023nerfdiff} and explicit methods~\cite{cao22fwd,guo2022fenvs,szymanowicz2023viewset}.
We use the deterministic variants of~\cite{gu2023nerfdiff,szymanowicz2023viewset} by using their reconstruction network in a single forward pass.
For CO3D we compare against PixelNeRF which we train for 400,000 iterations with their officially released code on the same data as used for our method.
Finally, on Objaverse-LVIS we compare to OpenLRM~\cite{he2023openlrm} (open-source version of LRM~\cite{hong24lrm:}): a large triplane-based reconstructor, trained on the full Objaverse dataset.
Since we are proposing a deterministic reconstruction method, we do not compare to methods that employ Score Distillation~\cite{liu23zero-1-to-3:,zhou2022sparsefusion} or feed-forward diffusion models~\cite{chan2023genvs,liu23one-2-3-45:,watson20223dim,xu24dmv3d:}.

Implementation details can be found in the supp.~mat.

\subsection{Evaluation of reconstruction quality}

\begin{table}[]
    \centering
    \setlength{\tabcolsep}{0.25pt}
    \footnotesize
    \begin{tabular}{l l c c c c c c }
       \toprule
        Method &  RC  & \multicolumn{3}{c}{1-view Cars} & \multicolumn{3}{c}{1-view Chairs}  \\
        {}     &  {} & PSNR $\uparrow$ & SSIM $\uparrow$ & LPIPS $\downarrow$ & PSNR $\uparrow$ & SSIM $\uparrow$ & LPIPS $\downarrow$ \\
       \midrule
       SRN                              & \xmark & 22.25 & 0.88 & 0.129 & 22.89 & 0.89 & 0.104 \\
       CodeNeRF                         & \xmark & 23.80 & 0.91 & 0.128 & 23.66 & 0.90 & 0.166 \\
       FE-NVS                           & \xmark & 22.83 & 0.91 & 0.099 & 23.21 & 0.92 & 0.077 \\
       ViewsetDiff w/o $\mathcal{D}$          & \xmark & 23.21 & 0.90 & 0.116 & 24.16 & 0.91 & 0.088 \\
       \midrule
       PixelNeRF                        & \checkmark & 23.17 & 0.89 & 0.146 & 23.72 & 0.90 & 0.128 \\
       VisionNeRF                       & \checkmark & 22.88 & 0.90 & 0.084 & 24.48 & 0.92 & 0.077 \\
       NeRFDiff w/o NGD                 & \checkmark & 23.95 & \textbf{0.92} & 0.092 & \textbf{24.80} & \textbf{0.93}& 0.070 \\
       \midrule
       \textbf{Ours}                    & \checkmark & \textbf{24.00} & \textbf{0.92} & \textbf{0.078} & 24.43 & \textbf{0.93} & \textbf{0.067} \\
       \bottomrule
    \end{tabular}
    \caption{\textbf{ShapeNet-SRN\@: Single-View Reconstruction.} Our method achieves State-of-the-Art reconstruction quality on all metrics on the Car dataset and on two metrics in the Chair dataset, while performing reconstruction in the camera view-space.
    `RC' indicates if a method can operate using only relative camera poses.}%
    \label{tab:single_view_reconstruction}
\end{table}

In line with related works~\cite{lin2023visionnerf,yu21pixelnerf:}, we assess the quality of the reconstructions by measuring novel view synthesis quality and report Peak Signal-to-Noise Ratio (PSNR), Structural Similarity (SSIM) and a perceptual loss (LPIPS).
We perform reconstruction from a given source view and render the 3D shape to unseen target views following standard protocols as detailed in the supp.~mat.

\subsubsection{Single-view 3D reconstruction}

\begin{figure}
    \centering
    \includegraphics[width=\columnwidth]{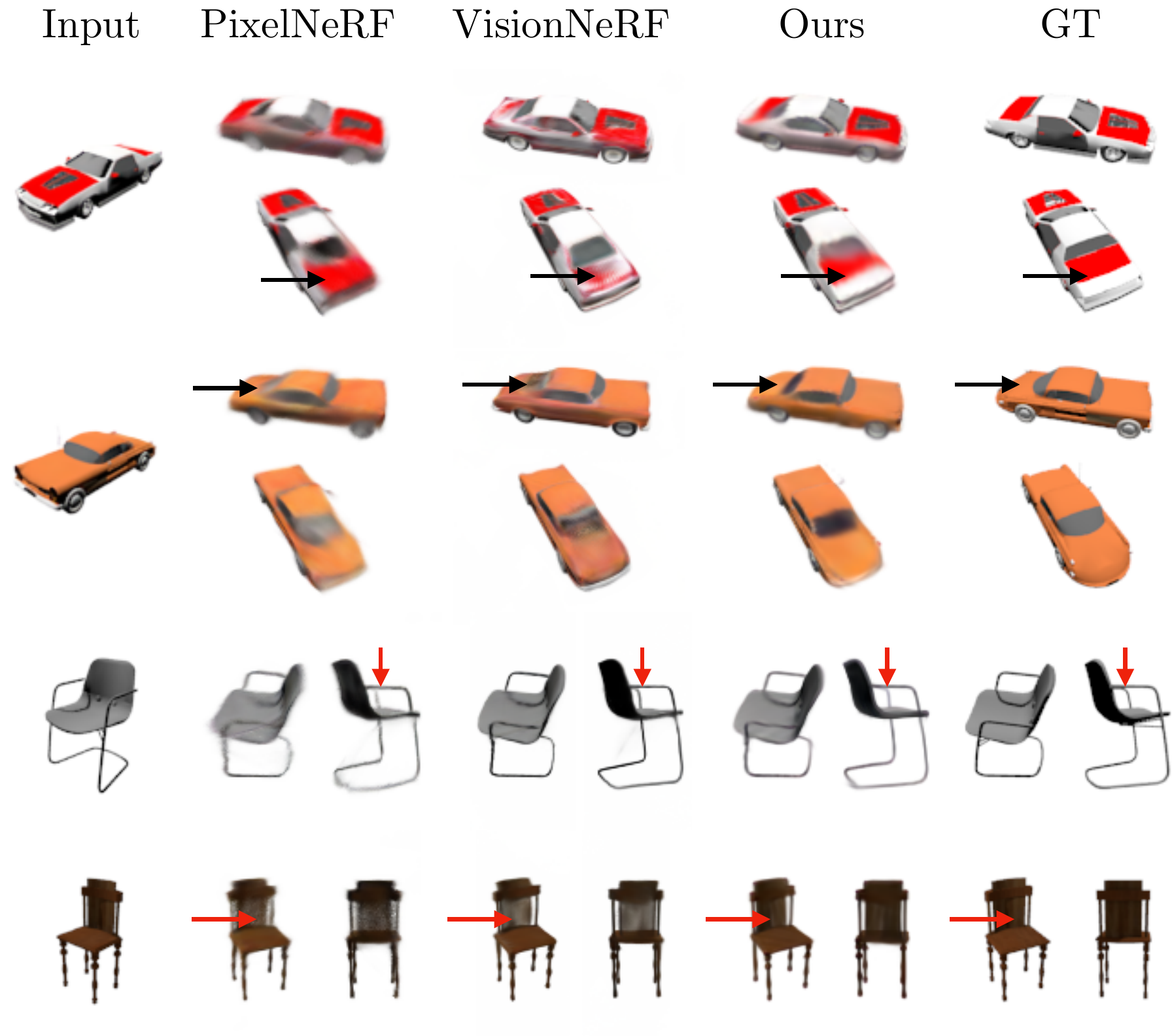}
    \caption{\textbf{ShapeNet-SRN Comparison.} Our method outputs more accurate reconstructions (cars' backs, top chair) and better represents thin regions (bottom chair).}%
    \label{fig:qualitative-shapenet-comparison}
\end{figure}

\begin{table}[]
    \centering
    \resizebox{0.99\columnwidth}{!}{%
    \begin{tabular}{l  l l l l l  l}
       \toprule
        {} & PixelNeRF~\cite{yu21pixelnerf:} & FE-NVS~\cite{guo2022fenvs} & FWD~\cite{cao22fwd} &  & VisionNeRF~\cite{lin2023visionnerf} & Ours  \\
       \midrule
        PSNR $\uparrow$ & 26.80 &  27.08 & 26.66 &  & 28.76 & \textbf{29.38} \\
        SSIM $\uparrow$ & 0.91 & 0.92 & 0.91 & & 0.93 & \textbf{0.95} \\
        LPIPS $\downarrow$ & 0.108 & 0.082 & 0.055 &  & 0.065 & \textbf{0.047} \\
       \bottomrule
    \end{tabular}
    }\vspace{-0.2cm}
    \caption{Our method achieves State-of-the-Art quality of single-view reconstruction on multi-class ShapeNet dataset.}%
    \label{tab:nmr}
\end{table}

\paragraph{ShapeNet.} 

In \cref{tab:single_view_reconstruction,fig:qualitative-shapenet-comparison} we compare the single-view reconstruction quality on the ShapeNet-SRN benchmark.
Our method outperforms all deterministic reconstructors in SSIM and LPIPS, obtaining sharper new views.
Furthermore, our method requires only relative camera poses instead of  absolute/canonical ones.
Qualitatively, our method does well in challenging situations with limited visibility and thin structures.
In \cref{tab:nmr}, we use instead the multi-category ShapeNet protocol we observe that our method outperforms more expensive baselines~\cite{lin2023visionnerf} across all metrics in the multi-category ShapeNet setting.

\paragraph{CO3D.}

\begin{figure}
    \centering
    \includegraphics[width=\columnwidth]{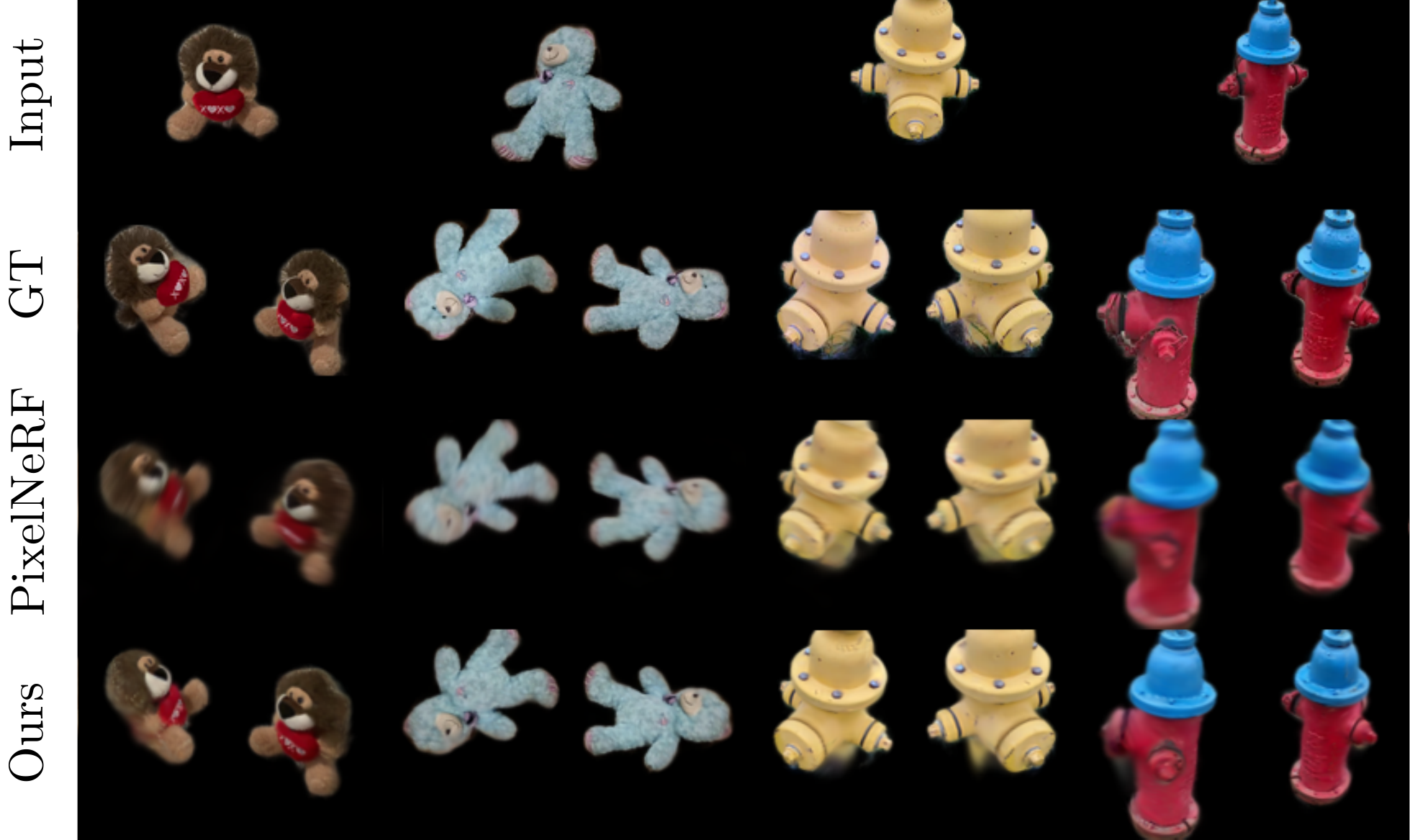}
    \caption{\textbf{CO3D Hydrants and Teddybears.} Our method outputs sharper reconstructions than PixelNeRF while being 100x faster in inference.}%
    \label{fig:co3d_comparison}
\end{figure}

On CO3D bears and hydrants, our model outperforms PixelNeRF on all metrics (\cref{tab:co3d_svr}), and qualitatively produces sharper images (\cref{fig:co3d_comparison}) while being 1,000$\times$ faster.

\begin{table}[]
    \centering
    \resizebox{0.97\columnwidth}{!}{%
    \begin{tabular}{c c c c c}
       \toprule
        Object & Method     & PSNR $\uparrow$ & SSIM $\uparrow$ & LPIPS $\downarrow$ \\
       \midrule
        Hydrant & PixelNeRF      & 21.76 & 0.78 & 0.203 \\ 
        Hydrant & \textbf{Ours}  & \textbf{21.80} & \textbf{0.80} & \textbf{0.150} \\ 
       \midrule
        Teddybear & PixelNeRF     & 19.38 & 0.65 & 0.290 \\ 
        Teddybear & \textbf{Ours} & \textbf{19.44} & \textbf{0.73} & \textbf{0.231} \\ 
       \bottomrule
    \end{tabular}
    }
    \caption{\textbf{CO3D\@: Single-View.} Our method outperforms PixelNeRF on this challenging benchmark across all metrics.}%
    \label{tab:co3d_svr}
\end{table}

\paragraph{Objaverse-LVIS and Google Scanned Objects.}

\begin{figure}
    \centering
    \includegraphics[width=0.94\columnwidth]{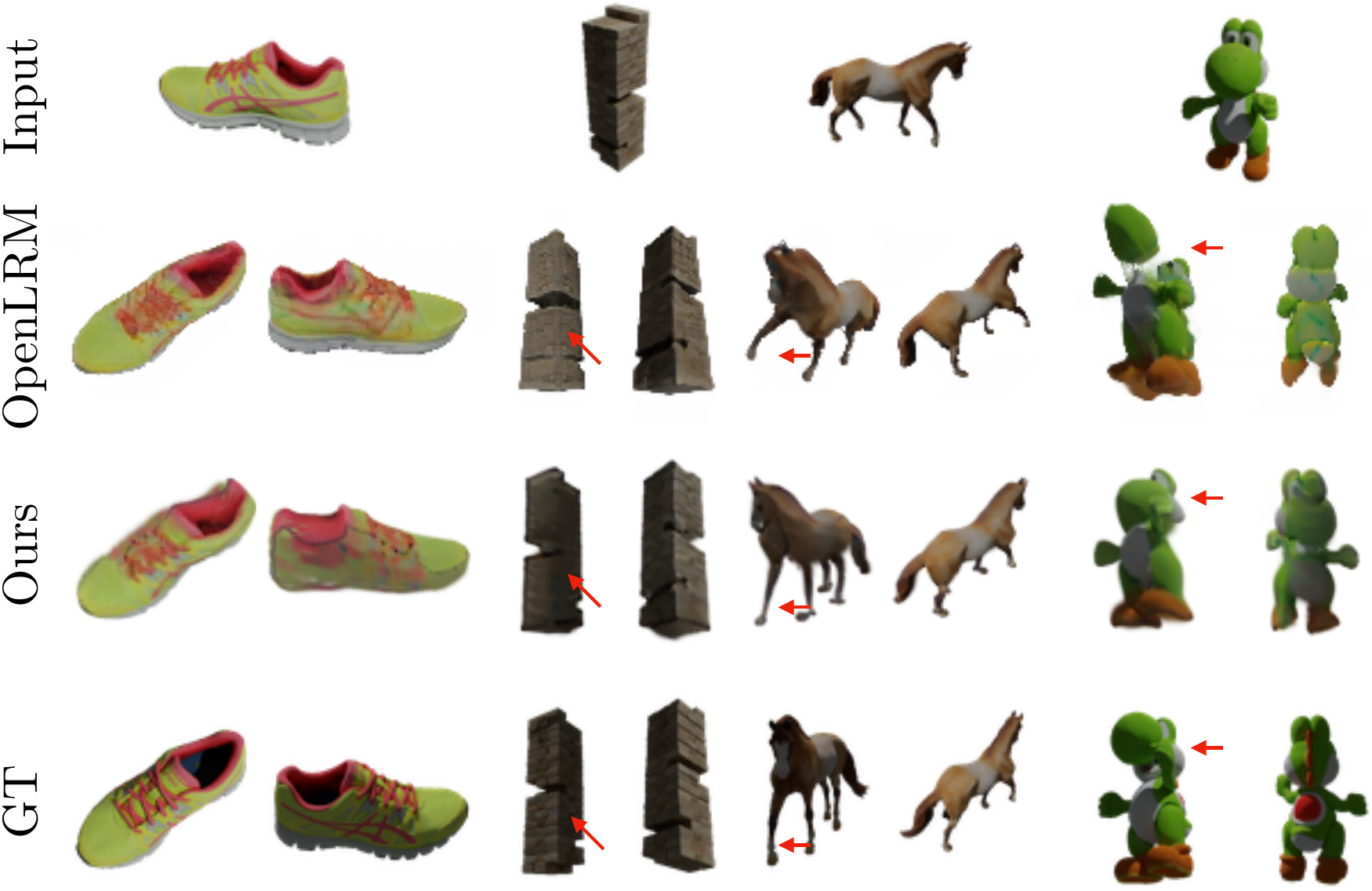}\vspace{-0.2cm}
    \caption{\textbf{Google Scanned Objects.} On large datasets our model has similar quality to much more expensive baselines (shoe). Our reconstructions have more accurate lighting (Jenga), object pose (horse) and shape (toy).}%
    \label{fig:gso_comparison}
\end{figure}

\begin{figure}
    \centering
    \includegraphics[width=0.94\columnwidth]{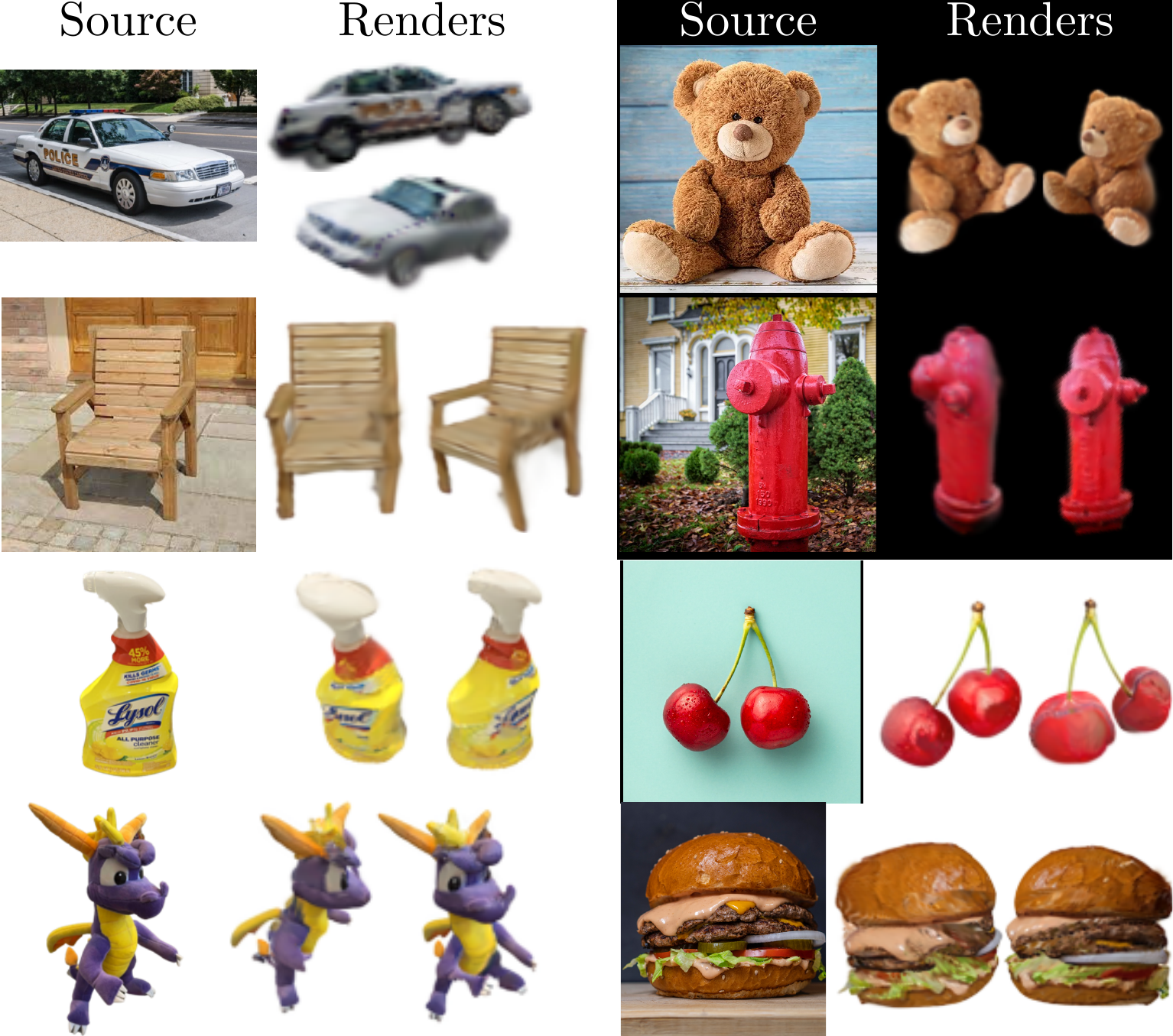}\vspace{-0.2cm}
    \caption{Our models trained on single classes (top) and on Objaverse (bottom) can be used on \textbf{in-the-wild} Internet images (right).}%
    \label{fig:itw_generalisation}
\end{figure}

We compare our method to OpenLRM~\cite{he2023openlrm}, an open-source version of the LRM model~\cite{hong24lrm:}, on the Google Scanned Objects~\cite{francis22gso} evaluation renderings from Free3D~\cite{zheng24free3d}.
Quantitatively, in \cref{tab:gso} \method outperforms OpenLRM and, qualitatively (\cref{fig:gso_comparison}), it is comparable. 
Our models perform well even on images collected from the Internet \cref{fig:itw_generalisation}, after removing backgrounds and resizing.
Remarkably, our method, trained for 7 GPU days, is able to compete with OpenLRM, which uses hundreds of GPUs for several days~\cite{he2023openlrm,hong24lrm:}.

\begin{table}[]
    \centering
    \begin{tabular}{c c c c c c}
       \toprule
        Method     & PSNR $\uparrow$ & SSIM $\uparrow$ & LPIPS $\downarrow$ \\
        \midrule
        OpenLRM    & 18.06 & 0.84 & 0.129 \\
        Ours    & \textbf{21.06} & \textbf{0.88} & \textbf{0.111} \\
       \bottomrule
    \end{tabular}
    \caption{\textbf{Google Scanned Objects: Single-View.} Our method outperforms the much more expensive LRM~\cite{he2023openlrm,hong24lrm:} on single-view open-world reconstruction.}%
    \label{tab:gso}
\end{table}

\subsubsection{Two-view 3D reconstruction}

We compare our multi-view reconstruction model on ShapeNet-SRN Cars by training it for two-view predictions (see \cref{tab:srn_cars_tvr}).
Prior work often relies on absolute camera pose conditioning, meaning that the model learns to rely on the canonical orientation of the object in the dataset.
This limits the applicability of these models, as in practice for a new image of an object, the absolute camera pose is of course unknown.
Here, only ours and PixelNeRF can deal with relative camera poses as input.
Interestingly, our method shows not only better performance than PixelNeRF in both real and synthetic data but also improves over SRN, CodeNeRF, and FE-NVS that rely on absolute camera poses.

\begin{table}[]
    \centering
    \resizebox{0.67\columnwidth}{!}{%
    \begin{tabular}{c c c c}
       \toprule
        Method & Relative & \multicolumn{2}{c}{2-view Cars}  \\
        {}     & Pose     & PSNR $\uparrow$ & SSIM $\uparrow$ \\
       \midrule
        SRN             & \xmark & 24.84 & 0.92 \\
        CodeNeRF        & \xmark & 25.71 & 0.91 \\
        FE-NVS          & \xmark & 24.64 & 0.93 \\
       \midrule
        PixelNeRF       & \checkmark & 25.66 & \textbf{0.94}  \\
        \textbf{Ours}   & \checkmark & \textbf{26.01} & \textbf{0.94} \\

       \bottomrule
    \end{tabular}
    }
    \caption{Two-view reconstruction on ShapeNet-SRN Cars.}%
    \vspace{-0.4cm}
    \label{tab:srn_cars_tvr}
\end{table}

\subsubsection{Ablations}

We evaluate the influence of individual components of our method, using a shorter training schedule than models in \cref{tab:single_view_reconstruction} for efficiency.
Ablations of the multi-view model are given in the supp.~mat.

We show the results of our ablation study for the single-view model in \cref{tab:ablations_svr}.
We train a model (w/o image) that uses a fully connected, unstructured output instead of a \method.
This model cannot transfer image information directly to their corresponding Gaussians and does not achieve good performance.
We also ablate predicting the depth along the ray by simply predicting 3D coordinates for each Gaussian.
This version also suffers from its inability to easily align the input image with the output.
Removing the 3D offset prediction mainly harms the backside of the object while leaving the front faces the same.
This results in a lower impact on the overall performance of this component.
Changing the degrees of freedom of appearance predictions (by fixing Gaussians to be isotropic or removing view-dependence) also reduced the image fidelity.
Finally, removing perceptual loss (w/o $\mathcal{L}_{\text{LPIPS}}$) results in a significant worsening of LPIPS, indicating this loss is important for perceptual sharpness of reconstructions.
Being able to use LPIPS in optimisation is a direct consequence of employing a fast-to-render representation and being able to render full images at training time.

\paragraph{Analysis.}

\begin{figure}
    \centering
    \includegraphics[width=\columnwidth]{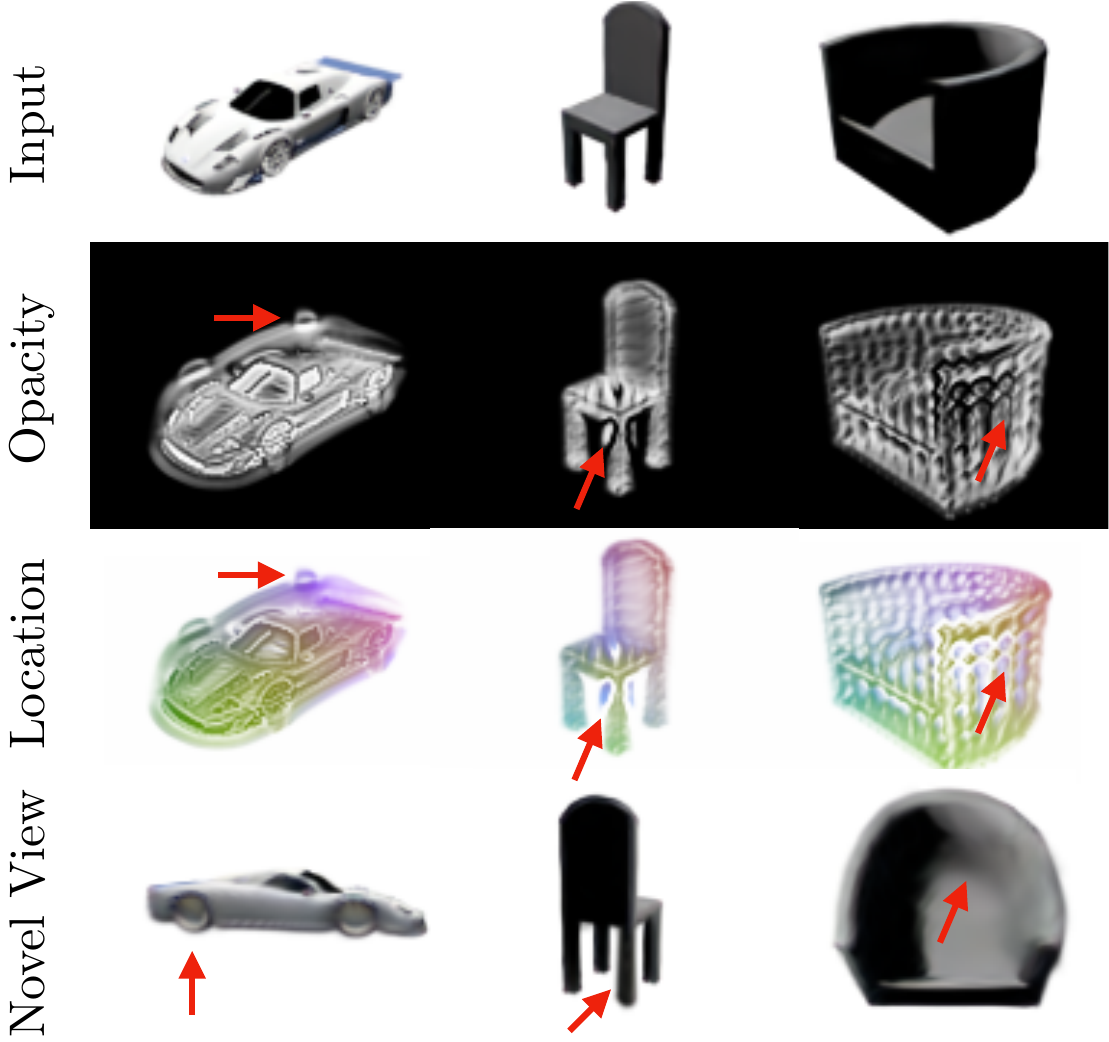}
    \caption{\textbf{Analysis.} Splatter Images represent full $360^{\circ}$ of objects by allocating background pixels to appropriate 3D locations (third row) to predict occluded elements like wheels (left) or chair legs (middle). Alternatively, it predicts offsets in the foreground pixels to represent occluded chair parts (right).}%
    \vspace{-0.4cm}
    \label{fig:splatter_image_analysis}
\end{figure}

In \cref{fig:splatter_image_analysis}, we analyse how 3D information is stored inside a \method.
Since all information is arranged in an image format, we can visualise each of the modalities: opacity, depth, and location.
Pixels of the input image that belong to the object tend to describe their corresponding 3D structure, while pixels outside of the object wrap around to close the object on the back.

\begin{table}[]
    \centering
    \begin{tabular}{l c c c c c}
        \toprule
        {}                          & PSNR $\uparrow$ & SSIM $\uparrow$ & LPIPS $\downarrow$ \\
        \midrule
        \textbf{Full model}         & \textbf{22.25} & \textbf{0.90} & \textbf{0.115} \\
        \midrule
        w/o image         & 20.60 & 0.87 & 0.152 \\

        w/o depth         & 21.21 & 0.88 & 0.145 \\
        w/o view dir.        & 21.77 & 0.89 & 0.121 \\
        isotropic            & 22.01 & 0.89 & 0.118 \\
        w/o offset        & 22.06 & \textbf{0.90} & 0.119 \\

        w/o $\mathcal{L}_{\text{LPIPS}}$         & 22.22 & 0.89 & 0.141 \\
       \bottomrule
    \end{tabular}
    \caption{\textbf{Ablations: Single-View Reconstruction.}}%
    \label{tab:ablations_svr}
\end{table}

\subsection{Evaluation of reconstruction efficiency}%
\label{s:efficiency}

\begin{table}[]
    \centering
    \resizebox{0.97\columnwidth}{!}{%
    \begin{tabular}{l c c c c c}
       \toprule
        {}     & RP & E $\downarrow$ & R $\downarrow$ & Forward $\downarrow$ & Test $\downarrow$ \\
       \midrule
        NeRFDiff                         & \checkmark & (0.031)        & (0.0180)        & (0.103)        & (4.531)        \\
        FE-NVS                           & \xmark     & (0.015)        & (0.0032)        & (0.028) & (0.815) \\
       \midrule
        VisionNeRF                       & \checkmark & 0.008          & 2.4312 & 9.733 & 607.8 \\
        PixelNeRF                        & \checkmark & \textbf{0.003} & 1.8572 & 7.432 & 463.3    \\
        ViewsetDiff                            & \xmark     & 0.025 & 0.0064 & 0.051 & 1.625 \\
       \midrule
        \textbf{Ours 2-view}             & \checkmark & 0.030          & \textbf{0.0017} & 0.037          & 0.455          \\
        \textbf{Ours 1-view}             & \checkmark & 0.026          & \textbf{0.0017} & \textbf{0.033} & \textbf{0.451} \\

       \bottomrule
    \end{tabular}
    }
    \caption{\textbf{Speed.} Time required for image encoding (E), rendering (R), the \emph{`Forward'} time, indicative of train-time efficiency and the \emph{`Test'} time, indicative of test-time efficiency. Our method is the most efficient in both train and test time across open-source available methods and only requires relative camera poses. `RP' indicates if a method can operate using only relative camera poses.}%
    \label{tab:speed}
\end{table}

A key advantage of the \method is its training and test time efficiency, which we assess below.

\paragraph{Test-time efficiency.}

First, we assess the \textit{`Test'} time speed, \ie, the time it takes for the trained model to reconstruct an object and generate a certain number of images.
We reference the evaluation protocol of the standard ShapeNet-SRN benchmark~\cite{sitzmann2019srns} and render 250 images at $128^2$ resolution.

Assessing wall-clock time fairly is challenging as it depends on many factors.
All measurements reported here are done on a single NVIDIA V100 GPU\@.
We use officially released code of Viewset Diffusion~\cite{szymanowicz2023viewset}, PixelNeRF~\cite{yu21pixelnerf:} and VisionNeRF~\cite{lin2023visionnerf} and rerun those on our hardware.
NeRFDiff~\cite{gu2023nerfdiff} and FE-NVS~\cite{guo2022fenvs} do not have code available, so we use their self-reported metrics.
According to the authors, FE-NVS was evaluated on the same type of GPU, while NeRFDiff does not include information about the hardware used and we were unable to obtain more information.
Since we could not perfectly control these experiments, the comparisons to NeRFDiff and FE-NVS are only indicative.
For Viewset Diffusion and NeRFDiff we report the time for a single pass through the reconstruction network.

\cref{tab:speed} reports the \textit{`Encoding'} (E) time, spent by the network to compute the object's 3D representation from an image, and the \textit{`Rendering'} (R) time, spent by the network to render new images from the 3D representation.
From those, we calculate the \textit{`Test'} time, equal to the \textit{`Encoding'} time plus 250 \textit{`Rendering'} time.
As shown in the last column of \cref{tab:speed}, our method is more than $1000\times$ faster in testing than PixelNeRF and VisionNeRF (while achieving equal or superior quality of reconstruction in \cref{tab:single_view_reconstruction}).
Our method is also faster than voxel-based Viewset Diffusion even though it does not require knowing the absolute camera pose.
The efficiency of our method is very useful to iterate quickly in research; for instance, evaluating our method on the full ShapeNet-Car validation set takes \textbf{less than 10 minutes} on a single GPU\@.
In contrast, PixelNeRF takes \textbf{45 GPU-hours}.

\paragraph{Train-time efficiency.}

Next, we assess the efficiency of the method during training.
Here, the encoding time becomes more significant because one typically renders only a few images to compute the reconstruction loss and obtain a gradient (\eg, because there are only so many views available in the training dataset, or because generating more views provides diminishing returns in terms of supervision).
As typical values (and as used by us in this work), we assume that the method is tasked with generating 4 new views at each iteration instead of 250 as before.
We call this the \textit{`Forward'} time and measure it the same way.
As shown in the \textit{`Forward'} column of \cref{tab:speed}, our method is $246\times$ faster at training time than implicit methods and $1.5\times$ than Viewset Diffusion, which uses an explicit representation.
With this, we can train models achieving state-of-the-art quality on a \textbf{single A6000 GPU} in 7 days, while VisionNeRF requires \textbf{16 A100 GPUs} for 5 days.
What is even more remarkable, we can train models on large datasets such as Objaverse on \textbf{two A6000 GPUs in 3.5 days}, while triplane-based methods such as LRM require \textbf{128 A100 GPUs for 3 days}~\cite{hong24lrm:}.
\section{Conclusion}%
\label{sec:conclusion}

We have presented \method, a simple method for single- or few-view 3D reconstruction.
The method uses an off-the-shelf 2D image-to-image network and predicts a pseudo-image containing one colored 3D Gaussian per pixel.
By combining fast inference with fast rendering via Gaussian Splatting, \method can be trained and evaluated quickly on synthetic and real benchmarks.
\method achieves state-of-the-art reconstruction performance without requiring absolute/canonical camera poses at test time, is simple to implement, and can be trained and tested much faster than many alternatives.

\paragraph{Ethics.}

We use various datasets in a manner compatible with their terms.
There is no processing of personal data.
For further details on ethics, data protection, and copyright please see \url{https://www.robots.ox.ac.uk/~vedaldi/research/union/ethics.html}.

\paragraph{Acknowledgements.}

S. Szymanowicz is supported by an EPSRC Doctoral Training Partnerships Scholarship (DTP) EP/R513295/1 and the Oxford-Ashton Scholarship.
A. Vedaldi is supported by ERC-CoG UNION 101001212.

{
    \small
    \bibliographystyle{ieeenat_fullname}
    \bibliography{main, vedaldi_general, vedaldi_specific}

\begin{thebibliography}{60}
\providecommand{\natexlab}[1]{#1}
\providecommand{\url}[1]{\texttt{#1}}
\expandafter\ifx\csname urlstyle\endcsname\relax
  \providecommand{\doi}[1]{doi: #1}\else
  \providecommand{\doi}{doi: \begingroup \urlstyle{rm}\Url}\fi

\bibitem[Anciukevicius et~al.(2022)Anciukevicius, Xu, Fisher, Henderson, Bilen, Mitra, and Guerrero]{anciukevicius22renderdiffusion:}
Titas Anciukevicius, Zexiang Xu, Matthew Fisher, Paul Henderson, Hakan Bilen, Niloy~J. Mitra, and Paul Guerrero.
\newblock {RenderDiffusion}: Image diffusion for 3d reconstruction, inpainting and generation.
\newblock In \emph{Proc. {CVPR}}, 2022.

\bibitem[Cao et~al.(2022)Cao, Rockwell, and Johnson]{cao22fwd}
Ang Cao, Chris Rockwell, and Justin Johnson.
\newblock Fwd: Real-time novel view synthesis with forward warping and depth.
\newblock In \emph{Proc. {CVPR}}, 2022.

\bibitem[Chan et~al.(2022)Chan, Lin, Chan, Nagano, Pan, Mello, Gallo, Guibas, Tremblay, Khamis, Karras, and Wetzstein]{chan22efficient}
Eric~R. Chan, Connor~Z. Lin, Matthew~A. Chan, Koki Nagano, Boxiao Pan, Shalini~De Mello, Orazio Gallo, Leonidas~J. Guibas, Jonathan Tremblay, Sameh Khamis, Tero Karras, and Gordon Wetzstein.
\newblock Efficient geometry-aware {3D} generative adversarial networks.
\newblock In \emph{Proc. {CVPR}}, 2022.

\bibitem[Chan et~al.(2023)Chan, Nagano, Chan, Bergman, Park, Levy, Aittala, Mello, Karras, and Wetzstein]{chan2023genvs}
Eric~R. Chan, Koki Nagano, Matthew~A. Chan, Alexander~W. Bergman, Jeong~Joon Park, Axel Levy, Miika Aittala, Shalini~De Mello, Tero Karras, and Gordon Wetzstein.
\newblock {GeNVS}: Generative novel view synthesis with {3D}-aware diffusion models.
\newblock In \emph{Proc. {ICCV}}, 2023.

\bibitem[Chang et~al.(2015)Chang, Funkhouser, Guibas, Hanrahan, Huang, Li, Savarese, Savva, Song, Su, Xiao, Yi, and Yu]{chang15shapenet}
Angel~X. Chang, Thomas~A. Funkhouser, Leonidas~J. Guibas, Pat Hanrahan, Qi{-}Xing Huang, Zimo Li, Silvio Savarese, Manolis Savva, Shuran Song, Hao Su, Jianxiong Xiao, Li Yi, and Fisher Yu.
\newblock {ShapeNet} an information-rich 3d model repository.
\newblock \emph{arXiv.cs}, abs/1512.03012, 2015.

\bibitem[Chen et~al.(2021)Chen, Xu, Zhao, Zhang, Xiang, Yu, and Su]{chen21mvsnerf:}
Anpei Chen, Zexiang Xu, Fuqiang Zhao, Xiaoshuai Zhang, Fanbo Xiang, Jingyi Yu, and Hao Su.
\newblock {MVSNeRF}: Fast generalizable radiance field reconstruction from multi-view stereo.
\newblock In \emph{Proc. {ICCV}}, 2021.

\bibitem[Chen et~al.(2022)Chen, Xu, Geiger, Yu, and Su]{chen22tensorf:}
Anpei Chen, Zexiang Xu, Andreas Geiger, Jingyi Yu, and Hao Su.
\newblock {TensoRF}: Tensorial radiance fields.
\newblock In \emph{arXiv}, 2022.

\bibitem[Chen et~al.(2023)Chen, Gu, Chen, Tian, Tu, Liu, and Su]{chen23ssdnerf}
Hansheng Chen, Jiatao Gu, Anpei Chen, Wei Tian, Zhuowen Tu, Lingjie Liu, and Hao Su.
\newblock Single-stage diffusion nerf: A unified approach to 3d generation and reconstruction.
\newblock In \emph{Proc. {ICCV}}, 2023.

\bibitem[Deitke et~al.(2023{\natexlab{a}})Deitke, Liu, Wallingford, Ngo, Michel, Kusupati, Fan, Laforte, Voleti, Gadre, VanderBilt, Kembhavi, Vondrick, Gkioxari, Ehsani, Schmidt, and Farhadi]{deitke23objaverse-xl:}
Matt Deitke, Ruoshi Liu, Matthew Wallingford, Huong Ngo, Oscar Michel, Aditya Kusupati, Alan Fan, Christian Laforte, Vikram Voleti, Samir~Yitzhak Gadre, Eli VanderBilt, Aniruddha Kembhavi, Carl Vondrick, Georgia Gkioxari, Kiana Ehsani, Ludwig Schmidt, and Ali Farhadi.
\newblock {Objaverse-XL}: {A} universe of {10M}+ {3D} objects.
\newblock \emph{CoRR}, abs/2307.05663, 2023{\natexlab{a}}.

\bibitem[Deitke et~al.(2023{\natexlab{b}})Deitke, Schwenk, Salvador, Weihs, Michel, VanderBilt, Schmidt, Ehsani, Kembhavi, and Farhadi]{deitke23objaverse:}
Matt Deitke, Dustin Schwenk, Jordi Salvador, Luca Weihs, Oscar Michel, Eli VanderBilt, Ludwig Schmidt, Kiana Ehsani, Aniruddha Kembhavi, and Ali Farhadi.
\newblock Objaverse: {A} universe of annotated {3D} objects.
\newblock In \emph{Proc. {CVPR}}, 2023{\natexlab{b}}.

\bibitem[Fan et~al.(2017)Fan, Su, and Guibas]{fan17point_set}
Haoqiang Fan, Hao Su, and Leonidas Guibas.
\newblock A point set generation network for 3d object reconstruction from a single image.
\newblock In \emph{Proc. {ICCV}}, 2017.

\bibitem[Francis et~al.(2022)Francis, Kinman, Reymann, Downs, Koenig, Hickman, McHugh, and Vanhoucke]{francis22gso}
Anthony~G. Francis, Brandon Kinman, Krista~Ann Reymann, Laura Downs, Nathan Koenig, Ryan~M. Hickman, Thomas~B. McHugh, and Vincent~Olivier Vanhoucke.
\newblock Google scanned objects: A high-quality dataset of 3d scanned household items.
\newblock In \emph{Proc. {ICRA}}, 2022.

\bibitem[Gu et~al.(2023)Gu, Trevithick, Lin, Susskind, Theobalt, Liu, and Ramamoorthi]{gu2023nerfdiff}
Jiatao Gu, Alex Trevithick, Kai-En Lin, Joshua~M Susskind, Christian Theobalt, Lingjie Liu, and Ravi Ramamoorthi.
\newblock Nerfdiff: Single-image view synthesis with nerf-guided distillation from 3d-aware diffusion.
\newblock In \emph{Proc. {ICML}}, 2023.

\bibitem[Guo et~al.(2022)Guo, Bautista, Colburn, Yang, Ulbricht, Susskind, and Shan]{guo2022fenvs}
Pengsheng Guo, Miguel~Angel Bautista, Alex Colburn, Liang Yang, Daniel Ulbricht, Joshua~M. Susskind, and Qi Shan.
\newblock Fast and explicit neural view synthesis.
\newblock In \emph{Proc. {WACV}}, 2022.

\bibitem[He and Wang(2023)]{he2023openlrm}
Zexin He and Tengfei Wang.
\newblock Openlrm: Open-source large reconstruction models.
\newblock \url{https://github.com/3DTopia/OpenLRM}, 2023.

\bibitem[Ho et~al.(2020)Ho, Jain, and Abbeel]{ho20denoising}
Jonathan Ho, Ajay Jain, and Pieter Abbeel.
\newblock Denoising diffusion probabilistic models.
\newblock In \emph{Proc. {NeurIPS}}, 2020.

\bibitem[Hong et~al.(2024)Hong, Zhang, Gu, Bi, Zhou, Liu, Liu, Sunkavalli, Bui, and Tan]{hong24lrm:}
Yicong Hong, Kai Zhang, Jiuxiang Gu, Sai Bi, Yang Zhou, Difan Liu, Feng Liu, Kalyan Sunkavalli, Trung Bui, and Hao Tan.
\newblock {LRM}: Large reconstruction model for single image to {3D}.
\newblock In \emph{Proc. {ICLR}}, 2024.

\bibitem[Jain et~al.(2021)Jain, Tancik, and Abbeel]{jain2021dietnerf}
Ajay Jain, Matthew Tancik, and Pieter Abbeel.
\newblock Putting nerf on a diet: Semantically consistent few-shot view synthesis.
\newblock In \emph{Proc. {ICCV}}, pages 5885--5894, 2021.

\bibitem[Jang and Agapito(2021)]{jang21codenerf:}
Wonbong Jang and Lourdes Agapito.
\newblock {CodeNeRF}: Disentangled neural radiance fields for object categories.
\newblock In \emph{Proc. {ICCV}}, 2021.

\bibitem[Karras et~al.(2022)Karras, Aittala, Aila, and Laine]{karras22elucidating}
Tero Karras, Miika Aittala, Timo Aila, and Samuli Laine.
\newblock Elucidating the design space of diffusion-based generative models.
\newblock In \emph{Proc. {NeurIPS}}, 2022.

\bibitem[Kato et~al.(2018)Kato, Ushiku, and Harada]{kato2018renderer}
Hiroharu Kato, Yoshitaka Ushiku, and Tatsuya Harada.
\newblock Neural 3d mesh renderer.
\newblock In \emph{Proc. {CVPR}}, 2018.

\bibitem[Kerbl et~al.(2023)Kerbl, Kopanas, Leimk{\"u}hler, and Drettakis]{kerbl233d-gaussian}
Bernhard Kerbl, Georgios Kopanas, Thomas Leimk{\"u}hler, and George Drettakis.
\newblock {3D Gaussian Splatting for Real-Time Radiance Field Rendering}.
\newblock \emph{Proc. {SIGGRAPH}}, 42\penalty0 (4), 2023.

\bibitem[Kingma and Ba(2014)]{kingma14adam:}
Diederik~P Kingma and Jimmy Ba.
\newblock Adam: A method for stochastic optimization.
\newblock \emph{arXiv preprint arXiv:1412.6980}, 2014.

\bibitem[Kulh{\'a}nek et~al.(2022)Kulh{\'a}nek, Derner, Sattler, and Babu{\v s}ka]{kulhanek22viewformer:}
Jon{\'a}{\v s} Kulh{\'a}nek, Erik Derner, Torsten Sattler, and Robert Babu{\v s}ka.
\newblock {ViewFormer}: {NeRF-free} neural rendering from few images using transformers.
\newblock In \emph{Proc. {ECCV}}, 2022.

\bibitem[Li et~al.(2024)Li, Tan, Zhang, Xu, Luan, Xu, Hong, Sunkavalli, Shakhnarovich, and Bi]{li24instant3d:}
Jiahao Li, Hao Tan, Kai Zhang, Zexiang Xu, Fujun Luan, Yinghao Xu, Yicong Hong, Kalyan Sunkavalli, Greg Shakhnarovich, and Sai Bi.
\newblock {Instant3D}: Fast text-to-{3D} with sparse-view generation and large reconstruction model.
\newblock \emph{Proc. {ICLR}}, 2024.

\bibitem[Lin et~al.(2023)Lin, Yen-Chen, Lai, Lin, Shih, and Ramamoorthi]{lin2023visionnerf}
Kai-En Lin, Lin Yen-Chen, Wei-Sheng Lai, Tsung-Yi Lin, Yi-Chang Shih, and Ravi Ramamoorthi.
\newblock Vision transformer for nerf-based view synthesis from a single input image.
\newblock In \emph{Proc. {WACV}}, 2023.

\bibitem[Liu et~al.(2021)Liu, Tucker, Jampani, Makadia, Snavely, and Kanazawa]{liu21infinite}
Andrew Liu, Richard Tucker, Varun Jampani, Ameesh Makadia, Noah Snavely, and Angjoo Kanazawa.
\newblock Infinite nature: Perpetual view generation of natural scenes from a single image.
\newblock In \emph{Proc. {ICCV}}, 2021.

\bibitem[Liu et~al.(2023{\natexlab{a}})Liu, Xu, Jin, Chen, T, Xu, and Su]{liu23one-2-3-45:}
Minghua Liu, Chao Xu, Haian Jin, Linghao Chen, Mukund~Varma T, Zexiang Xu, and Hao Su.
\newblock One-2-3-45: Any single image to {3D} mesh in 45 seconds without per-shape optimization.
\newblock In \emph{Proc. {NeurIPS}}, 2023{\natexlab{a}}.

\bibitem[Liu et~al.(2023{\natexlab{b}})Liu, Wu, Hoorick, Tokmakov, Zakharov, and Vondrick]{liu23zero-1-to-3:}
Ruoshi Liu, Rundi Wu, Basile~Van Hoorick, Pavel Tokmakov, Sergey Zakharov, and Carl Vondrick.
\newblock Zero-1-to-3: Zero-shot one image to {3D} object.
\newblock In \emph{Proc. {ICCV}}, 2023{\natexlab{b}}.

\bibitem[Liu et~al.(2024)Liu, Lin, Zeng, Long, Liu, Komura, and Wang]{liu2023syncdreamer}
Yuan Liu, Cheng Lin, Zijiao Zeng, Xiaoxiao Long, Lingjie Liu, Taku Komura, and Wenping Wang.
\newblock Syncdreamer: Learning to generate multiview-consistent images from a single-view image.
\newblock \emph{Proc. {ICLR}}, 2024.

\bibitem[Long et~al.(2022)Long, Lin, Wang, Komura, and Wang]{long22sparseneus:}
Xiaoxiao Long, Cheng Lin, Peng Wang, Taku Komura, and Wenping Wang.
\newblock {SparseNeuS}: Fast generalizable neural surface reconstruction from sparse views.
\newblock In \emph{Proc. {ECCV}}, 2022.

\bibitem[Melas-Kyriazi et~al.(2023{\natexlab{a}})Melas-Kyriazi, Rupprecht, Laina, and Vedaldi]{melas2023pc2}
Luke Melas-Kyriazi, Christian Rupprecht, Iro Laina, and Andrea Vedaldi.
\newblock {PC}2: Projection-conditioned point cloud diffusion for single-image 3d reconstruction.
\newblock In \emph{Proc. {CVPR}}, 2023{\natexlab{a}}.

\bibitem[Melas-Kyriazi et~al.(2023{\natexlab{b}})Melas-Kyriazi, Rupprecht, Laina, and Vedaldi]{melas2023realfusion}
Luke Melas-Kyriazi, Christian Rupprecht, Iro Laina, and Andrea Vedaldi.
\newblock Realfusion: 360° reconstruction of any object from a single image.
\newblock In \emph{Proc. {CVPR}}, 2023{\natexlab{b}}.

\bibitem[Mildenhall et~al.(2020)Mildenhall, Srinivasan, Tancik, Barron, Ramamoorthi, and Ng]{mildenhall20nerf:}
Ben Mildenhall, Pratul~P. Srinivasan, Matthew Tancik, Jonathan~T. Barron, Ravi Ramamoorthi, and Ren Ng.
\newblock {NeRF}: Representing scenes as neural radiance fields for view synthesis.
\newblock In \emph{Proc. {ECCV}}, 2020.

\bibitem[M{\"{u}}ller et~al.(2023)M{\"{u}}ller, Siddiqui, Porzi, Bul{\`{o}}, Kontschieder, and Nie{\ss}ner]{muller23diffrf:}
Norman M{\"{u}}ller, Yawar Siddiqui, Lorenzo Porzi, Samuel~Rota Bul{\`{o}}, Peter Kontschieder, and Matthias Nie{\ss}ner.
\newblock {DiffRF}: Rendering-guided {3D} radiance field diffusion.
\newblock In \emph{Proc. {CVPR}}, 2023.

\bibitem[Perez et~al.(2018)Perez, Strub, de~Vries, Dumoulin, and Courville]{perez18film:}
Ethan Perez, Florian Strub, Harm de Vries, Vincent Dumoulin, and Aaron~C. Courville.
\newblock {FiLM}: Visual reasoning with a general conditioning layer.
\newblock In \emph{AAAI}, 2018.

\bibitem[Qi et~al.(2017)Qi, Su, Mo, and Guibas]{qi2017pointnet}
Charles~R Qi, Hao Su, Kaichun Mo, and Leonidas~J Guibas.
\newblock Pointnet: Deep learning on point sets for 3d classification and segmentation.
\newblock In \emph{Proc. {CVPR}}, 2017.

\bibitem[Reizenstein et~al.(2021)Reizenstein, Shapovalov, Henzler, Sbordone, Labatut, and Novotny]{reizenstein21co3d}
Jeremy Reizenstein, Roman Shapovalov, Philipp Henzler, Luca Sbordone, Patrick Labatut, and David Novotny.
\newblock Common objects in 3d: Large-scale learning and evaluation of real-life 3d category reconstruction.
\newblock In \emph{Proc. {ICCV}}, 2021.

\bibitem[Rematas et~al.(2021)Rematas, Martin{-}Brualla, and Ferrari]{rematas21sharf:}
Konstantinos Rematas, Ricardo Martin{-}Brualla, and Vittorio Ferrari.
\newblock {ShaRF}: Shape-conditioned radiance fields from a single view.
\newblock In \emph{Proc. {ICML}}, 2021.

\bibitem[Rockwell et~al.(2021)Rockwell, Fouhey, and Johnson]{rockwell21pixelsynth}
Chris Rockwell, David~F. Fouhey, and Justin Johnson.
\newblock Pixelsynth: Generating a 3d-consistent experience from a single image.
\newblock In \emph{Proc. {ICCV}}, 2021.

\bibitem[Rombach et~al.(2022)Rombach, Blattmann, Lorenz, Esser, and Ommer]{rombach22high-resolution}
Robin Rombach, Andreas Blattmann, Dominik Lorenz, Patrick Esser, and Bj{\"{o}}rn Ommer.
\newblock High-resolution image synthesis with latent diffusion models.
\newblock In \emph{Proc. {CVPR}}, 2022.

\bibitem[Ronneberger et~al.(2015)Ronneberger, Fischer, and Brox]{ronneberger15u-net:}
Olaf Ronneberger, Philipp Fischer, and Thomas Brox.
\newblock U-net: Convolutional networks for biomedical image segmentation.
\newblock In \emph{Proc. {MICCAI}}, 2015.

\bibitem[R{\"u}ckert et~al.(2022)R{\"u}ckert, Franke, and Stamminger]{ruckert22adop}
Darius R{\"u}ckert, Linus Franke, and Marc Stamminger.
\newblock Adop: Approximate differentiable one-pixel point rendering.
\newblock In \emph{ACM Trans. on Graphics (TOG)}, 2022.

\bibitem[Shi et~al.(2023)Shi, Wang, Ye, Mai, Li, and Yang]{shi2023MVDream}
Yichun Shi, Peng Wang, Jianglong Ye, Long Mai, Kejie Li, and Xiao Yang.
\newblock Mvdream: Multi-view diffusion for 3d generation.
\newblock \emph{arXiv:2308.16512}, 2023.

\bibitem[Sitzmann et~al.(2019)Sitzmann, Zollh{\"o}fer, and Wetzstein]{sitzmann2019srns}
Vincent Sitzmann, Michael Zollh{\"o}fer, and Gordon Wetzstein.
\newblock Scene representation networks: Continuous 3d-structure-aware neural scene representations.
\newblock In \emph{Proc. {NeurIPS}}, 2019.

\bibitem[Song et~al.(2021)Song, Meng, and Ermon]{song2020denoising}
Jiaming Song, Chenlin Meng, and Stefano Ermon.
\newblock Denoising diffusion implicit models.
\newblock \emph{Proc. {ICLR}}, 2021.

\bibitem[Sun et~al.(2022)Sun, Sun, and Chen]{sun22direct}
Cheng Sun, Min Sun, and Hwann{-}Tzong Chen.
\newblock Direct voxel grid optimization: Super-fast convergence for radiance fields reconstruction.
\newblock In \emph{Proc. {CVPR}}, 2022.

\bibitem[Szymanowicz et~al.(2023)Szymanowicz, Rupprecht, and Vedaldi]{szymanowicz2023viewset}
Stanislaw Szymanowicz, Christian Rupprecht, and Andrea Vedaldi.
\newblock Viewset diffusion: (0-)image-conditioned {3D} generative models from {2D} data.
\newblock In \emph{Proc. {ICCV}}, 2023.

\bibitem[Tewari et~al.(2023)Tewari, Yin, Cazenavette, Rezchikov, Tenenbaum, Durand, Freeman, and Sitzmann]{tewari2023diffusion}
Ayush Tewari, Tianwei Yin, George Cazenavette, Semon Rezchikov, Joshua~B. Tenenbaum, Frédo Durand, William~T. Freeman, and Vincent Sitzmann.
\newblock Diffusion with forward models: Solving stochastic inverse problems without direct supervision.
\newblock In \emph{Proceedings of Advances in Neural Information Processing Systems (NeurIPS)}, 2023.

\bibitem[Tochilkin et~al.(2024)Tochilkin, Pankratz, Liu, Huang, Letts, Li, Liang, Laforte, Jampani, and Cao]{tochilkin24triposr:}
Dmitry Tochilkin, David Pankratz, Zexiang Liu, Zixuan Huang, Adam Letts, Yangguang Li, Ding Liang, Christian Laforte, Varun Jampani, and Yan-Pei Cao.
\newblock {TripoSR:} fast {3D} object reconstruction from a single image.
\newblock 2403.02151, 2024.

\bibitem[Wang et~al.(2021)Wang, Wang, Genova, Srinivasan, Zhou, Barron, Martin{-}Brualla, Snavely, and Funkhouser]{wang21ibrnet:}
Qianqian Wang, Zhicheng Wang, Kyle Genova, Pratul~P. Srinivasan, Howard Zhou, Jonathan~T. Barron, Ricardo Martin{-}Brualla, Noah Snavely, and Thomas~A. Funkhouser.
\newblock Ibrnet: Learning multi-view image-based rendering.
\newblock In \emph{Proc. {CVPR}}, 2021.

\bibitem[Watson et~al.(2023)Watson, Chan, Martin-Brualla, Ho, Tagliasacchi, and Norouzi]{watson20223dim}
Daniel Watson, William Chan, Ricardo Martin-Brualla, Jonathan Ho, Andrea Tagliasacchi, and Mohammad Norouzi.
\newblock Novel view synthesis with diffusion models.
\newblock In \emph{Proc. {ICLR}}, 2023.

\bibitem[Wiles et~al.(2020)Wiles, Gkioxari, Szeliski, and Johnson]{wiles20synsin}
Olivia Wiles, Georgia Gkioxari, Richard Szeliski, and Justin Johnson.
\newblock Synsin: End-to-end view synthesis from a single image.
\newblock In \emph{Proc. {CVPR}}, 2020.

\bibitem[Xu et~al.(2024)Xu, Tan, Luan, Bi, Wang, Li, Shi, Sunkavalli, Wetzstein, Xu, and Zhang]{xu24dmv3d:}
Yinghao Xu, Hao Tan, Fujun Luan, Sai Bi, Peng Wang, Jiahao Li, Zifan Shi, Kalyan Sunkavalli, Gordon Wetzstein, Zexiang Xu, and Kai Zhang.
\newblock {DMV3D}: Denoising multi-view diffusion using {3D} large reconstruction model.
\newblock In \emph{Proc. {ICLR}}, 2024.

\bibitem[Yu et~al.(2021)Yu, Ye, Tancik, and Kanazawa]{yu21pixelnerf:}
Alex Yu, Vickie Ye, Matthew Tancik, and Angjoo Kanazawa.
\newblock {PixelNeRF}: Neural radiance fields from one or few images.
\newblock In \emph{Proc. {CVPR}}, 2021.

\bibitem[Zhang et~al.(2018)Zhang, Isola, Efros, Shechtman, and Wang]{zhang2018perceptual}
Richard Zhang, Phillip Isola, Alexei~A Efros, Eli Shechtman, and Oliver Wang.
\newblock The unreasonable effectiveness of deep features as a perceptual metric.
\newblock In \emph{Proc. {CVPR}}, 2018.

\bibitem[Zheng and Vedaldi(2024)]{zheng24free3d}
Chuanxia Zheng and Andrea Vedaldi.
\newblock Free3d: Consistent novel view synthesis without 3d representation.
\newblock In \emph{Proc. {CVPR}}, 2024.

\bibitem[Zhou et~al.(2021)Zhou, Du, and Wu]{zhou21pvd}
Linqi Zhou, Yilun Du, and Jiajun Wu.
\newblock 3d shape generation and completion through point-voxel diffusion.
\newblock In \emph{Proc. {ICCV}}, 2021.

\bibitem[Zhou and Tulsiani(2023)]{zhou2022sparsefusion}
Zhizhuo Zhou and Shubham Tulsiani.
\newblock Sparsefusion: Distilling view-conditioned diffusion for 3d reconstruction.
\newblock In \emph{Proc. {CVPR}}, 2023.

\bibitem[Zwicker et~al.(2001)Zwicker, Pfister, van Baar, and Gross]{zwicker01ewa-volume}
Matthias Zwicker, Hanspeter Pfister, Jeroen van Baar, and Markus~H. Gross.
\newblock {EWA} volume splatting.
\newblock In \emph{Proc. {IEEE} Visualization Conference,}, 2001.

\end{thebibliography}
}

\appendix
\clearpage
\setcounter{page}{1}
\maketitlesupplementary

\section{Additional results}

\begin{figure*}
    \centering
    \includegraphics[width=0.8\textwidth]{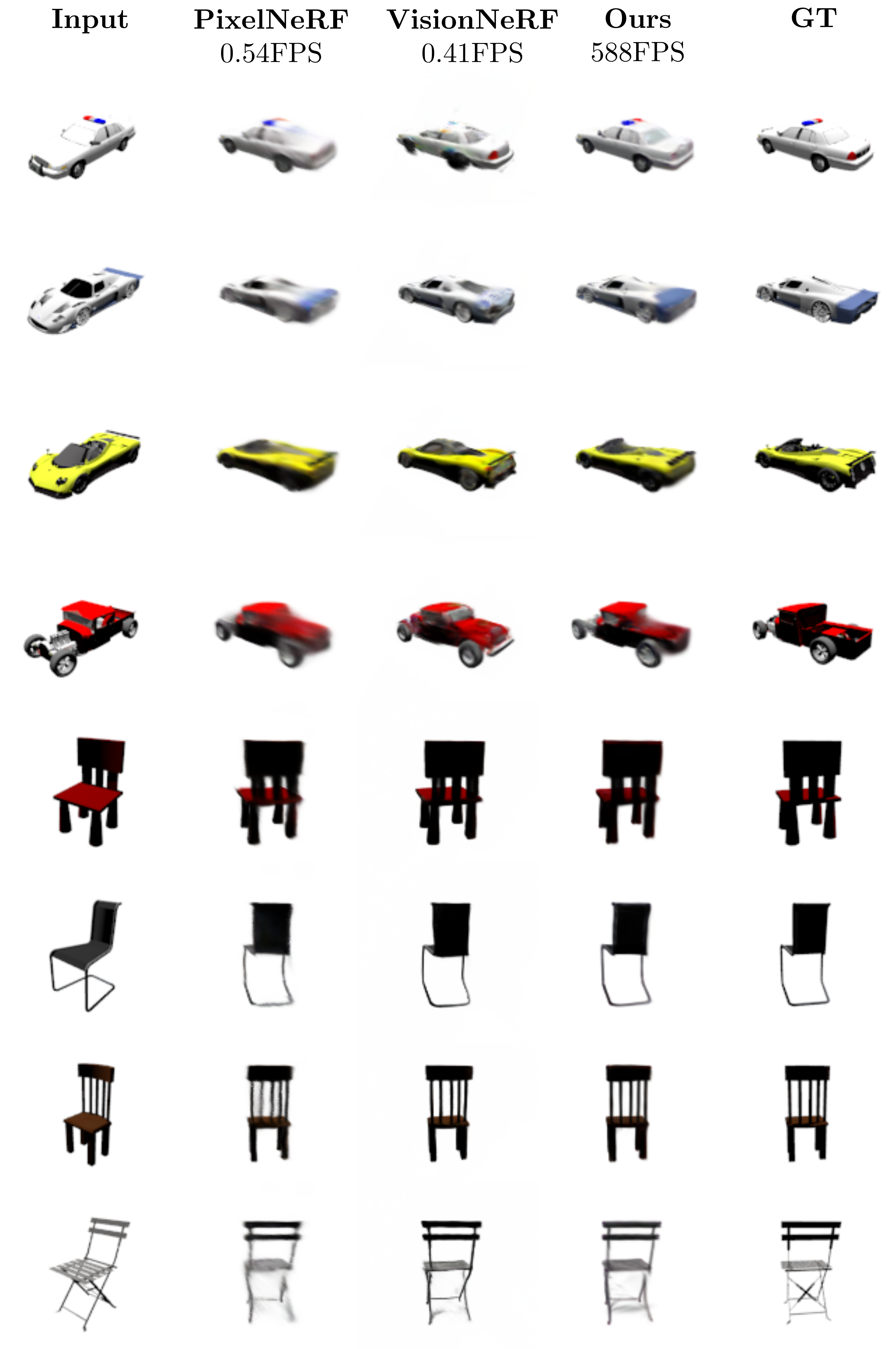}
    \caption{\textbf{ShapeNet-SRN.} Our method (fourth column) outputs reconstructions which are better than PixelNeRF (second column) and more or equally accurate than VisionNeRF (third column) while rendering 3 orders of magnitude faster (rendering speed in Frames Per Second denoted underneath method name).}%
    \label{fig:shapenet_comparison_supp}
\end{figure*}

\begin{figure*}
    \centering
    \includegraphics[width=0.58\textwidth]{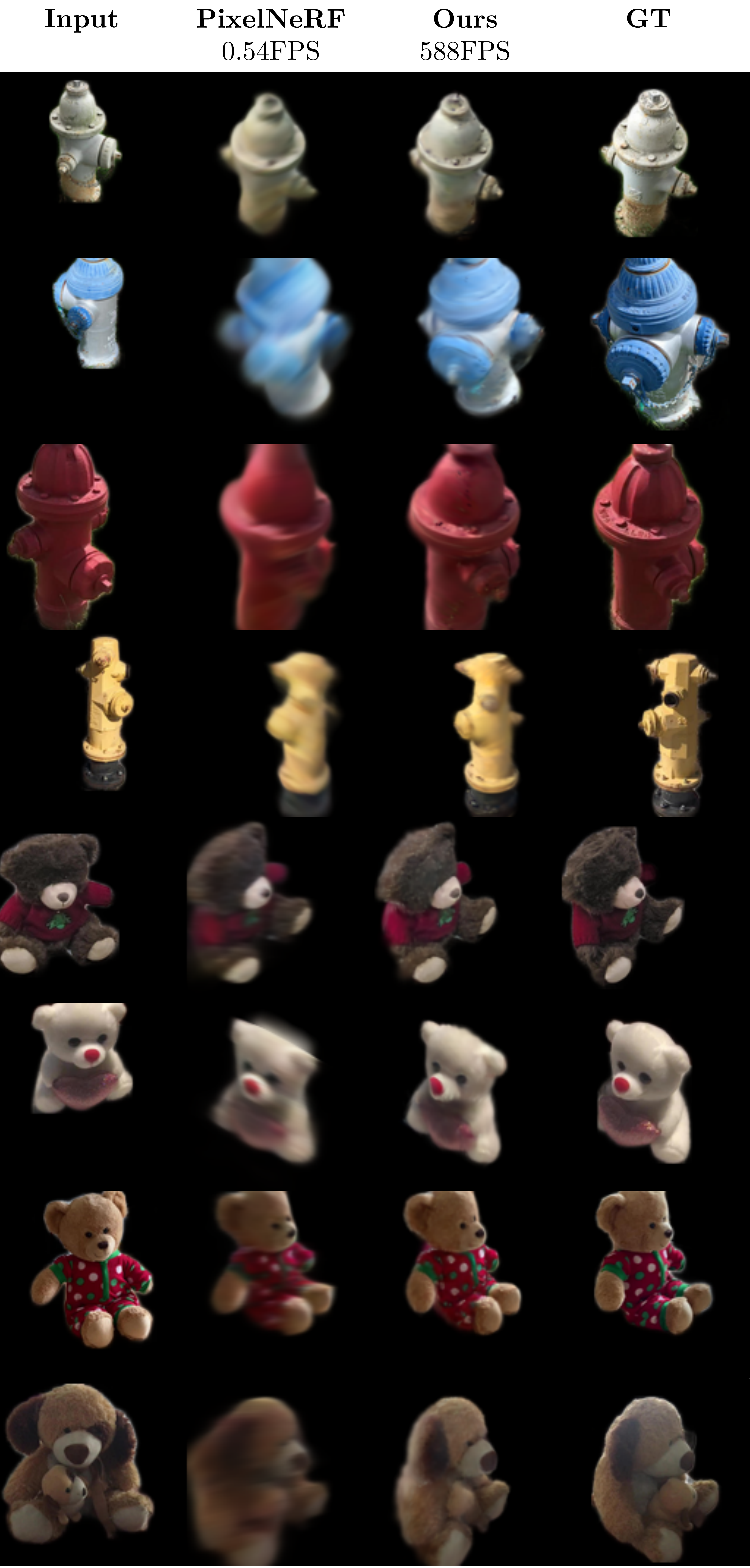}
    \caption{\textbf{CO3D.} Our method (third column) outputs reconstructions which are sharper than PixelNeRF (second column) while rendering 3 orders of magnitude faster (rendering speed in Frames Per Second denoted underneath method name).}%
    \label{fig:co3d_comparison_supp}
\end{figure*}

\begin{figure*}
    \centering
    \includegraphics[width=0.58\textwidth]{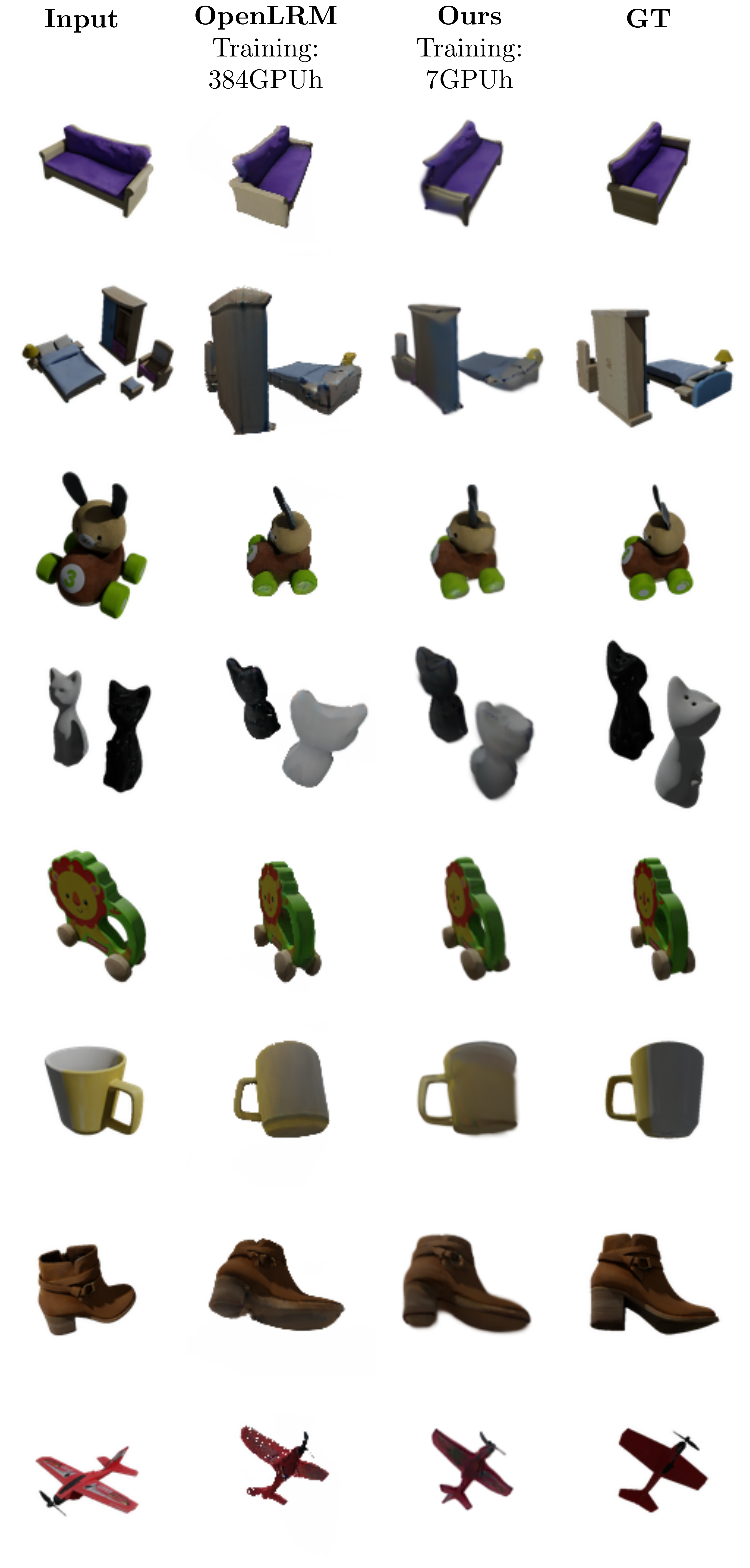}
    \caption{\textbf{Google Scanned Objects.} Our method (third column) outputs reconstructions which are comparable in quality to OpenLRM (second column) while requiring $\times 50$ less resources to train.}%
    \label{fig:gso_comparison_supp}
\end{figure*}

\paragraph{Additional qualitative results} Our project website contains a short summary of Splatter Image, videos of comparisons of our method to baselines and additional results from our method on the 4 object classes and the 2 multi-class datasets.
Moreover, we present static comparisons of our method to PixelNeRF~\cite{yu21pixelnerf:} and VisionNeRF on ShapeNet-SRN Cars and Chairs in~\cref{fig:shapenet_comparison_supp}, as well as static comparisons of our method to PixelNeRF on CO3D Hydrants and Teddybears in~\cref{fig:co3d_comparison_supp}.
In~\cref{fig:gso_comparison_supp} we present additional static comparisons of our method to OpenLRM on the Google Scanned Objects dataset.

\paragraph{Multi-view model ablation.} 
\Cref{tab:ablations_mvr} ablates the multi-view model.
We individually remove the multi-view attention blocks, the camera embedding and the warping component of the multi-view model and find that they all are important to achieve the final performance.

\begin{table}[]
    \centering
    \begin{tabular}{l c c c c c}
        \toprule
        {}                     & PSNR $\uparrow$ & SSIM $\uparrow$ & LPIPS $\downarrow$ \\
        \midrule
        \textbf{Full model}             & \textbf{24.11} &\textbf{0.92} & \textbf{0.087} \\
        \midrule
        w/o cross-view attn    & 23.68 & \textbf{0.92} & 0.091 \\
        w/o cam embed          & 23.91 & \textbf{0.92} & 0.088 \\
        w/o warping            & 23.84 & \textbf{0.92} & 0.088 \\
       \bottomrule
    \end{tabular}
    \caption{\textbf{Ablations: Multi-View Reconstruction.}}%
    \label{tab:ablations_mvr}
\end{table}

\section{Data details}

\subsection{ShapeNet-SRN Cars and Chairs}
We follow standard protocol in the ShapeNet-SRN datasets.
We use the images, camera intrinsics, camera poses and data splits as provided by the dataset~\cite{sitzmann2019srns} at $128 \times 128$ resolution and train our method using \textit{relative} camera poses: the reconstruction is done in the view space of the conditioning camera.
For single-view reconstruction, we use view 64 as the conditioning view and in two-view reconstruction we use views 64 and 128 as conditioning.
All other available views are used as target views in which we compute novel view synthesis merics. 

\subsection{CO3D}\label{s:preprocessing}

We use the first frame as input and all other frames as target frames.
We use all testing sequences in the Hydrant and Teddybear classes where the first conditioning frame has a valid foreground mask (with probability $p>0.8$).
In practice, this means evaluating on 49 `Hydrant' and 93 `Teddybear' sequences.

\paragraph{Image center-cropping.}

Similarly to recent methods~\cite{chan2023genvs,tewari2023diffusion} we take the largest crop in the original images centered on the principal point and resize to $128 \times 128$ resolution with Lanczos interpolation.
Similarly to many single- and few-view reconstruction methods~\cite{kulhanek22viewformer:,yu21pixelnerf:,zhou2022sparsefusion} we also remove backgrounds.
We adjust the focal length accordingly with the resulting transformations.
This is the only pre-processing we do -- CO3D objects already have their point clouds normalised to zero-mean and unit variance.

\paragraph{Predicting Gaussian positions.}

Estimating the distance between the object and the camera from visual information alone is a challenging problem in this dataset: focal lengths vary between and within sequences, objects are partially cropped, and global scene parameters such as distance to the object, camera trajectory and the angle at which objects are viewed all vary, posing a challenge to both our and baseline methods.
Thus, for both PixelNeRF and our method we set the center of prediction to the center of the object.

In our method we achieve this by setting $z_\text{near} = z_{\text{gt}} - w$ and $z_\text{far} = z_{\text{gt}} + w$, where $z_{\text{gt}}$ is the ground truth distance from the object to the source camera and $w$ is a fixed scalar $w=2.0$.
In PixelNeRF, we provide the network with $x = x_v - z_{\text{gt}}$ where $x$ is the sample location at which we query the network and $x_v$ is the sample location in camera view space.
$z_{\text{gt}}$ is computed as the perpendicular distance (along camera z-axis) to the world origin, which coincides with the center of the point cloud in CO3D.

\subsection{Multi-class ShapeNet.}
Identically to prior work, we use images, splits and camera parameters from NMR~\cite{kato2018renderer} which provides $64 \times 64$ renders from cameras at fixed elevations.
For direct comparison with prior work~\cite{lin2023visionnerf,yu21pixelnerf:} we use the same source and target views for evaluation.

\subsection{Objaverse and GSO data details.}\label{s:objaverse_data}
We use renders from Zero-1-to-3~\cite{liu23zero-1-to-3:}, filtered by the objects which appear in the LVIS subset to use only high-quality assets. 
The data is rendered at $512 \times 512$ resolution with focal length $560 \text{px}$ with cameras pointing at the center of the object at randomly sampled distances.
We resize data to $128 \times 128$ resolution with Lanczos interpolation, adjusting the focal length accordingly.
At training and testing time we rescale the ground truth camera positions so that the distance from the object to the camera is a fixed scalar $d=2$.
GSO renders provided by Free3D~\cite{zheng24free3d} were rendered with the same parameters (resolution, distances, focal length) and we apply the same resolution scaling, focal length adjustment and camera scale adjustment at evaluation time.

\section{Implementation details.}

\subsection{Splatter Image training.}

We train our model (based on SongUNet~\cite{song2020denoising}) with $\mathcal{L}_{2}$ reconstruction loss (Eq.4 main paper) on 3 unseen views and the conditioning view for 800,000 iterations.
We use the network implementation from~\cite{karras22elucidating}.
For single-class models, we use the Adam optimizer~\cite{kingma14adam:} with learning rate $5 \times 10^{-5}$ and batch size $8$.
For multi-class ShapeNet model we use the same learning rate and batch size $32$.
Batch sizes are mainly dictated by GPU memory limits.
For rasterization, we use the Gaussian Splatting implementation of~\cite{kerbl233d-gaussian}.
After 800,000 iterations we decrease the learning rate by a factor of $10$ and train for a further 100,000 (Cars, Hydrants, Teddybears), 150,000 (multi-class ShapeNet) or 200,000 (Chairs) iterations with the loss $\mathcal{L} = (1-\alpha) \mathcal{L}_{2} + \alpha \mathcal{L}_{\text{LPIPS}}$ and $\alpha=0.01$.
Training done is on a single NVIDIA A6000 GPU and takes around 7 days.

\paragraph{Large dataset training.} Training on Objaverse is done with Mixed Precision and effective batch size $32$.
We train first for $350,000$ iterations with learning rate $5 \times 10^{-5}$ and $\alpha=0$, followed by $40,000$ iterations with learning rate $6.3 \times 10^{-5}$ and $\alpha=0.338$.
Training takes place on two NVIDIA A6000 GPUs for around $3.5$ days.

\paragraph{Regularizers.} For CO3D we additionally use regularisation losses to prevent exceedingly large or vanishingly small Gaussians for numerical stability.
We regularize large Gaussians with the mean of their activated scale $s=\exp \hat{s}$ when it is bigger than a threshold scale $s_{\text{big}} = 20$.

$\mathcal{L}_{\text{big}} = (\sum_i s_i \mathbbm{1}(s_i>s_{\text{big}})) / (\sum_i \mathbbm{1}(s_i>s_{\text{big}}))$.

Small Gaussians are regularized with a mean of their negative deactivated scale $\hat{s}$ when it is smaller than a threshold $\hat s_{\text{small}} = -5$:
$
\mathcal{L}_{\text{small}}
=
(\sum_i -\hat s_i \mathbbm{1}(\hat s_i<\hat s_{\text{smal
l}}))
/
(\sum_i \mathbbm{1}(\hat s_i< \hat s_{\text{small}})).
$

\paragraph{Ablations.} Due to computational costs, ablation models are trained at a shorter schedule 100k iterations with $\mathcal{L}_{2}$ and further 25k with $\mathcal{L}_{2}$ and $\mathcal{L}_{\text{LPIPS}}$ with $\alpha=0.1$.

\subsection{PixelNeRF.}

For ShapeNet (single-class and multi-class) we use the scores reported in the original paper~\cite{yu21pixelnerf:}, as we train and evaluate on the same data.
For training on CO3D, we use the official PixelNeRF implementation~\cite{yu21pixelnerf:}.
We use the same preprocessed data as for our method.
We modify the activation function of opacity from ReLU to Softplus with the $\beta$ parameter $\beta=3.0$ for improved training stability.
Parametrization of the sampling points to be centered about the ground truth distance to the camera $z_\text{gt}$ as discussed in~\cref{s:preprocessing} is available as default in the official implementation.
As in original work, we train for $400,000$ iterations.

\subsection{OpenLRM.}
OpenLRM was trained assuming distance to the object $d=1.9$ and field-of-view $FOV=40^{\circ}$.
To match this, we rescale the ground truth cameras so that the source camera was at distance $d=1.9$ from the object.
For exact comparison we use the same data for the baselines as for our method.
For a fair comparison, we pass the $128 \times 128$ image as an input and render novel views at $128 \times 128$ too.
Through experimentation we found that the best quantitative results were achieved by assuming the same field-of-view as at training time $FOV=40^{\circ}$.

\section{Training resource estimate}

We compare the compute resources needed at training time by noting the GPU used, its capacity, the number of GPUs and the number of days needed for training in~\cref{tab:compute_resources}.
We report the compute resources reported in original works, where available.
NeRFDiff only reports the resources needed to train their `Base' models and the authors did not respond to our clarification emails about their `Large' models which we compare against in the main paper.
We thus report an estimate of such resources which we obtained by multiplying the number of GPUs used in the `Base' models by a factor of $2$.
Our method is significantly cheaper than VisionNeRF and NeRFDiff.
The resources required are similar to those of Viewset Diffusion and PixelNeRF, while we achieve better performance and do not require absolute camera poses.
The difference between our method and prior works is even more striking on large datasets like Objaverse, where our method is $\times 50$ cheaper than LRM.

\begin{table}[]
    \centering
    \resizebox{0.97\columnwidth}{!}{%
    \begin{tabular}{c c c c c c}
       \toprule
        Method & GPU & Memory & $\#$ GPUs & Days & GPU $\times$ Days \\
       \midrule
        VisionNeRF                          & A100 & 80G & 16 & 5 & 80 \\
        NeRFDiff                            & A100 & 80G & 16* & 3 & 48 \\
        ViewDiff                            & A40 & 48G & 2 & 3 & 6 \\
        PixelNeRF                           & TiRTX & 24G & 1 & 6 & 6 \\
        \midrule
        \textbf{Ours - small scale}                     & A6000 & 48G & 1 & 7 & 7 \\
        \midrule
        LRM / OpenLRM*                            & A100 & 40G & 128 & 3 & 384 \\
        \textbf{Ours - Objaverse}                       & A6000 & 48G & 2 & 3.5 & 7 \\
       \midrule
    \end{tabular}
    }
    \caption{\textbf{Training resources}. Ours, Viewset Diffusion and PixelNeRF have significantly lower compute costs than VisionNeRF and NeRFDiff. Our method is $\times 50$ cheaper to train than LRM. Memory denotes the memory capacity of the GPU. * denotes estimates.}%
    \label{tab:compute_resources}
\end{table}

\section{Covariance warping implementation}

As described in Sec.~3.4 in the main paper, the 3D Gaussians are warped from one view's reference frame to another with $\tilde \Sigma = R \Sigma R^\top$ where R is the relative rotation matrix of the reference frame transformation.
The covariance is predicted using a 3-dimensional scale and quaternion rotation so that $\Sigma = R_{q}SR_{q}^\top$ where $S = \mathrm{diag}\left(\exp( \hat s )\right)^{2}$.
Thus the warping is applied by applying rotation matrix $R$ to the orientation of the Gaussian $\tilde R_{q} = R R_{q}$.
In practice this is implemented in the quaternion space with the composition of the predicted quaternion $q$ and the quaternion representation of the relative rotation $p = m2q(R)$ where $m2q$ denotes the matrix-to-quaternion transformation, resulting in $\tilde q = p q$.

\end{document}